\documentclass[11pt,a4paper]{article}

\usepackage[T1]{fontenc}
\usepackage{lmodern}
\usepackage[margin=2.5cm]{geometry}
\usepackage{amsmath,amssymb,amsthm}
\usepackage{booktabs}
\usepackage{graphicx}
\usepackage{enumitem}
\usepackage{xcolor}
\usepackage{float}
\usepackage[colorlinks=true,linkcolor=blue!50!black,citecolor=blue!50!black,urlcolor=blue!50!black]{hyperref}

\newtheorem{proposition}{Proposition}
\newtheorem{remark}{Remark}

\newcommand{\oplusm}{\oplus}
\newcommand{\otimesm}{\otimes}
\newcommand{\eps}{\varepsilon}

\title{Learning in the Tropics:\\
Max-Plus Algebra as a Fast Simulator for Reinforcement Learning\\[1mm]
\large with an Application to Job-Shop Scheduling}
\author{Boussad ADDAD}
\date{November 22, 2026}

\begin{document}
\maketitle

\begin{abstract}
Reinforcement learning (RL) needs enormous numbers of simulated trajectories, so for discrete event systems (DES) the simulator is often the bottleneck: event-queue and fixed-step engines are general, but slow and hard to vectorize. We observe that a large class of DES, those governed by synchronization and delays, is \emph{linear in the max-plus algebra} when resources are not shared (timed event graphs), and \emph{switched} max-plus linear once resource-sharing conflicts are resolved (networked conflicting timed event graphs, NCTEGs). A simulation step is then a short sequence of additions and maxima over a sparse matrix: exact, free of event queue and clock, and easy to compile. We propose to use such models as the environment of RL: the Markov decision process has the exact max-plus recursion as its transition function, the state can be taken relative to a reference date (translation equivariance), and a sequential conflict-resolution scheme guarantees live (deadlock-free) modes. On job shops from $3\times3$ to $20\times10$, a compiled sparse max-plus simulator returns the same makespans as our event-driven and time-driven references and is up to $194\times$ and $419\times$ faster, with a 2.6 to 6.0 times lower peak Python heap; most of this factor comes from compilation and sparse storage, which the max-plus form makes easy, not from the algebra alone ($1.9\times$--$4.5\times$ uncompiled).
As an application, a hybrid policy (a softmax over the operations competing for each machine and a Gaussian head for release dates) is trained by policy gradient on repetitive job shops with due-date tracking. It lowers the mean cost by 24--62\% relative to five dispatching rules and a one-round-lookahead MPC, and by 18--26\% relative to a derivative-free search in the same policy class. A stronger baseline, a one-round MPC with an active-schedule candidate pool and a terminal cost, matches the policy on $3\times3$ shops and beats it by 13--37\% on $6\times4$ and $10\times5$ shops, but needs 150--190 times more computation per decision. Combining the two, by sampling a hundred schedules per round from the learned policy and choosing among them with the max-plus model, gives a mean cost 8--34\% below that MPC in five of six settings (the bootstrap interval excludes zero in four) at about a third of its decision time. Compiling policy and simulator with Numba leaves every result unchanged and makes training 6 to 35 times faster than the NumPy implementation; inside the training loop the simulator itself takes only 15--27\% of the episode time. Job shop scheduling is one application: the contribution is the pairing of RL with max-plus simulators, which should apply wherever the dynamics are (switched) max-plus linear.
\end{abstract}


\section{Introduction}

Reinforcement learning (RL) trades data for generality: a policy is learned from a very large number of simulated trajectories, because interacting with the real system is slow, costly or unsafe. The cost of the simulator therefore bounds what can be learned in a given time. For continuous control, much of the recent progress comes from making simulation itself cheaper, with compiled and massively parallel physics engines \cite{makoviychuk2021,freeman2021}. Discrete event systems (DES), such as factories, transport networks, data pipelines or supply chains, have no comparable engine. They are usually simulated with an event queue or with a fixed time step, in branching, general-purpose code that is hard to vectorize and whose cost grows with the number of events or with the length of the horizon.

\paragraph{Idea.}\footnote{The core idea behind this work, combining max-plus algebra with reinforcement learning (RL), dates back to 2019. A photograph of the original handwritten draft is included in the Appendix. The author lacked the time and opportunity to pursue it then, so the idea sat untouched until November 7, 2026, when the author came across the draft by chance. By then, AI had become capable enough to make implementing it feasible in a few days of testing and tuning. The first version was developed with Claude Sonnet 5.2 over three days (November 7, 8 and 11, 2026), then reviewed and refined with Claude Opus 5.5 on November 22, 2026.}
Many DES have a structure that makes them much cheaper to simulate than the general case: their behaviour is driven by \emph{synchronization} (an activity waits for several others to finish) and \emph{delays}. When no resource is shared, such systems are modelled by timed event graphs (TEGs), whose firing dates obey linear equations in the max-plus semiring \cite{baccelli1992}. When resources are shared (a machine, a robot or a track section used by several flows), the order in which the flows use the resource is a decision; networked conflicting timed event graphs (NCTEGs) model this, and once every conflict is resolved the evolution is a switched max-plus linear (SMPL) system \cite{addad2012,vandenboom2006}. In both cases the next dates are obtained from the previous ones by a few additions and maxima. There is no event list to maintain, no clock to advance, and no time discretization: the model is exact. We propose to use this representation as the \emph{environment} of reinforcement learning, and, because a step is so cheap, also as the \emph{model} with which a learned policy can compare candidate decisions at run time.

\paragraph{Why this has not been enough so far.} Max-plus algebra has a mature control theory, based on residuation, just-in-time control and model matching \cite{hardouin2018,baccelli1992}, and max-plus linear and SMPL models can be controlled by model predictive control \cite{deschutter2001,vandenboom2006,vandenboom2020}. These approaches use the model analytically and optimize over a horizon; the number of binary variables grows with the horizon and the number of conflicts, the program is re-solved at every step, and uncertain durations are not handled naturally. A learned policy moves this cost offline, but only if simulations are cheap, which is precisely what the max-plus structure offers.

\paragraph{Contributions.}
\begin{enumerate}[leftmargin=*]
\item \emph{The idea and its formalization.} We cast control of max-plus (switched) systems as a Markov decision process whose transition function is the exact max-plus recursion, and we prove two structural properties that matter for learning: translation equivariance, which justifies a relative-date state, and liveness of a sequential conflict-resolution scheme (Sections~\ref{sec:model} and~\ref{sec:app}).
\item \emph{A simulation benchmark.} We compare three max-plus implementations (NumPy dense, Numba dense, Numba sparse) with an event-driven and a time-driven simulator on job shops from $3\times3$ to $20\times10$. Outputs are identical; the Numba sparse implementation is up to $194\times$ faster than the event-driven and $419\times$ faster than the time-driven simulator, with lower memory use, and we separate the effect of compilation from that of the max-plus structure (Section~\ref{sec:bench}).
\item \emph{An application.} We train a hybrid policy, which resolves machine conflicts and sets release dates, by REINFORCE on repetitive job shops with due-date tracking, with and without duration noise. We compare it with dispatching rules, two one-round MPC controllers (one of them non-myopic), and a derivative-free search in the same policy class; we combine it with the max-plus model at decision time; we report an ablation that isolates the release dates, the computational cost of NumPy and Numba implementations, and the limits of the evidence (Section~\ref{sec:app}).
\item \emph{A scope statement.} We say precisely which systems fit the approach and which do not, and which claims are supported by our experiments and which are conjectures (Sections~\ref{sec:scope} and~\ref{sec:disc}).
\end{enumerate}

\section{Related Work}

\paragraph{Simulation throughput in RL.} Accelerating the environment is a major lever in RL: GPU-based physics simulation and differentiable or compiled engines have brought large reductions in training time for continuous control \cite{makoviychuk2021,freeman2021}. Our work follows the same logic for event-driven systems, using the algebraic structure of the dynamics instead of parallel hardware. Just-in-time compilation of numerical Python code is provided by Numba \cite{lam2015}.

\paragraph{Max-plus control of TEGs.} The algebraic theory is given in \cite{baccelli1992,heidergott2006}; control and state estimation are surveyed in \cite{hardouin2018}. Most results are open-loop or residuation-based feedback laws for systems without conflicts. Model predictive control of max-plus linear systems was introduced in \cite{deschutter2001}.

\paragraph{Conflicts and resource sharing.} NCTEGs and their time-varying max-plus representation, with a job-shop application, were introduced in \cite{addad2012}. Control of NCTEGs under strict time constraints, using switching max-plus models and control places, was studied in \cite{aberkane2021,aberkane2022}. Optimal control of timed event graphs sharing resources, with output-reference updates, is addressed in \cite{schafaschek}. These works design controllers with guarantees for the constraints considered; to our knowledge none learns a policy by reinforcement learning.

\paragraph{Switched max-plus linear systems and scheduling.} SMPL models and model predictive control of semi-cyclic systems were developed in \cite{vandenboom2006,vandenboom2020}; switched max-plus linear-dual inequalities for multi-product processing networks in \cite{zorzenon2022}; tracking of dynamic deadlines with uncontrollable workloads in SMPL systems in \cite{deadlines2024}. These approaches rely on optimization over a horizon. The repetitive job shop studied here is close to the cyclic (recurrent) job shop of the scheduling literature \cite{hanen1994}, which optimizes a periodic schedule rather than tracking due dates round by round.

\paragraph{Learning for scheduling.} Deep RL for job-shop scheduling learns priority dispatching rules end to end \cite{zhang2020}, often with graph neural networks \cite{song2022,hgt2026}. Closest in spirit to our use of a formal DES model as environment, \cite{petrirl2024} trains event-based RL on a Petri-net model of the job shop and \cite{hyper2026} uses a timed colored Petri net as environment; the max-plus recursion is the algebraic counterpart of such Petri-net environments for the subclass of event graphs with resolved conflicts. Max-plus techniques have also been used inside reinforcement learning as value-function approximators \cite{fqi2024}, which is a different use from ours: here max-plus algebra describes the controlled system and serves as the simulator, not the value function.

\paragraph{Position.} In the searches we performed we did not find prior work that trains a reinforcement learning policy for the joint choice of input dates and conflict resolutions on NCTEG/SMPL models, nor one that benchmarks max-plus recursions as RL simulators against event-driven and time-driven engines.

\section{Max-Plus Models as RL Environments}\label{sec:model}

\subsection{Max-plus algebra}
The max-plus semiring has addition $a\oplusm b=\max(a,b)$ and multiplication $a\otimesm b=a+b$, with zero element $\eps=-\infty$ and unit $0$. For matrices, $(A\otimesm x)_i=\max_j(A_{ij}+x_j)$. A TEG whose places contain at most one token has dates satisfying $x(n+1)=A\otimesm x(n)\oplusm B\otimesm u(n+1)$ and $y(n+1)=C\otimesm x(n+1)$.

\subsection{NCTEG as a switched max-plus linear system}
An NCTEG is a set of TEGs connected by conflict places. Let $d_n$ collect the conflict resolutions of round $n$. For fixed $d_n$ the net restricted to the chosen arcs is a TEG. Separating token-free places (same-round dependencies, matrix $A_0$) from one-token places (matrix $A_1$) gives
\begin{equation}
x(n+1)=A_0(d_n)\otimesm x(n+1)\;\oplusm\;A_1\otimesm x(n)\;\oplusm\;B\otimesm u(n+1),
\qquad y(n+1)=C\otimesm x(n+1),
\label{eq:smpl}
\end{equation}
whose smallest solution is $x(n+1)=A_0(d_n)^{*}\otimesm\big(A_1\otimesm x(n)\oplusm B\otimesm u(n+1)\big)$, where $^{*}$ is the Kleene star. It exists when $A_0(d)$ has no circuit of positive weight, i.e.\ when the net has no token-free circuit (liveness). This is the modelling route of \cite{addad2012,aberkane2021} written in the notation of switching max-plus linear systems \cite{vandenboom2006}.

\subsection{Structural properties}
\begin{proposition}[Translation equivariance]\label{prop:equiv}
For every real $c$, if $(x,u,d)$ produces $y$ under \eqref{eq:smpl}, then $(x+c,u+c,d)$ produces $y+c$.
\end{proposition}
\begin{proof}
In max-plus algebra $A\otimesm(x+c)=(A\otimesm x)+c$, and the same holds for $B\otimesm(u+c)$ and for the Kleene star, which is a finite maximum of sums of entries of $A_0$ applied to its argument. Every term of the recursion is therefore shifted by $c$.
\end{proof}
If the targets are shifted by the same constant, the tracking error, and hence the cost, is unchanged. A policy observing dates relative to a reference date loses no optimality and applies at any absolute time. In RL terms, the state can be expressed relative to a reference date, which removes the unbounded drift of absolute dates and makes the learned policy independent of the time origin.

\subsection{The environment as a Markov decision process}\label{sec:mdp}
Let the round index $n$ play the role of time. A round is one cycle of the system (for a manufacturing line, one batch of every job); its duration in the real world is irrelevant to the simulator. The MDP is defined as follows.
\begin{itemize}[leftmargin=*]
\item \emph{State:} the dates $x(n)$, preferably relative to a reference date $r_n$ (Proposition~\ref{prop:equiv}), together with the targets or exogenous data of the next rounds.
\item \emph{Action:} the continuous input dates $u(n)$ and the discrete conflict resolutions $d_n$ (which determine the mode of \eqref{eq:smpl}).
\item \emph{Transition:} $x(n+1)=A_0(d_n)^{*}\otimesm\big(A_1\otimesm x(n)\oplusm B\otimesm u(n+1)\big)$; deterministic, or stochastic when the entries of $A_0,A_1$ (durations) are random.
\item \emph{Reward:} any function of the outputs $y(n+1)=C\otimesm x(n+1)$ and of the targets, for example a tracking cost.
\end{itemize}
The transition is the model itself: there is no modelling error between ``simulator'' and ``model'', and the simulator can be compared exactly to the event-driven semantics (Section~\ref{sec:bench}). In practice the discrete action $d_n$ is usually too large to be chosen at once; it is built by a sequence of decisions inside the round, one per conflict, each conditioned on the partial schedule (Section~\ref{sec:method}). The MDP above is the round-level view, and the policy factorizes over these intra-round decisions.

\subsection{Why a max-plus step is cheap}
Fix a mode $d$. When the net has no token-free circuit, $A_0(d)$ is nilpotent and its arcs can be ordered topologically. The smallest solution of \eqref{eq:smpl} is then obtained by a single pass over the arcs in that order, with one addition and one maximum per arc, and without ever forming $A_0(d)^{*}$.

\begin{remark}[Cost of one round]\label{rem:cost}
With a topological order, one round costs $O(|\mathcal{A}|)$ elementary operations, where $|\mathcal{A}|$ is the number of arcs of the net. For a job shop with $J$ jobs and $M$ machines, $|\mathcal{A}|=O(JM)$ while the dense matrices $A_0,A_1$ have $(JM)^2$ entries. An event-driven simulator needs, in addition, an event list with $O(\log)$ insertions and removals per event, and a time-driven simulator performs one step per time unit (about $\mathrm{makespan}/dt$ steps), whatever the number of events. A dense implementation has a cost per round that grows with the square of the number of operations, a sparse one only with the number of arcs.
\end{remark}
This remark explains where speedup can come from; it is not a proof of the observed factors, which also depend on the implementation language and are measured in Section~\ref{sec:bench}. The three max-plus implementations compared there are: (i) dense matrices in NumPy; (ii) dense matrices compiled with Numba; (iii) a sparse arc list processed in topological order, compiled with Numba. A second use of the cheap step is \emph{batch evaluation}: many candidate modes $d$ for the same state can be evaluated at once, which a model-based controller or a learned policy can use to compare decisions at run time (Section~\ref{sec:method}).

\subsection{Scope: which systems fit}\label{sec:scope}
The approach applies when, for a given resolution of the conflicts, the dates of the events obey a max-plus linear recursion, that is, when each event occurs after the \emph{latest} of a set of other events plus a delay. This covers timed event graphs and, through the mode, systems with resource sharing and routing choices, including random durations (random matrix entries). Examples of application areas studied in the max-plus literature are manufacturing lines and job shops, cyclic timetables in rail and metro networks \cite{vandenboom2006}, and processing networks \cite{zorzenon2022}; we only test job shops. The approach does not cover dynamics that are not of this form, such as preemption, durations that depend on the system load in a non-max-plus way, or continuous dynamics coupled to the events; for such systems the max-plus model would be an approximation or would not apply.

\section{Speed of Max-Plus Simulators}\label{sec:bench}

\subsection{Setup}
\begin{figure}[t]
\centering
\includegraphics[width=\textwidth]{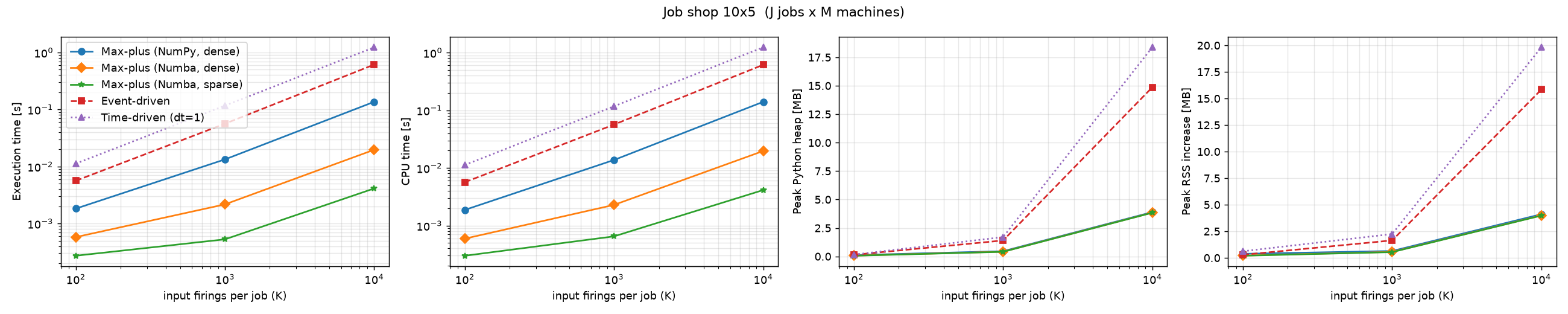}\\[1mm]
\includegraphics[width=\textwidth]{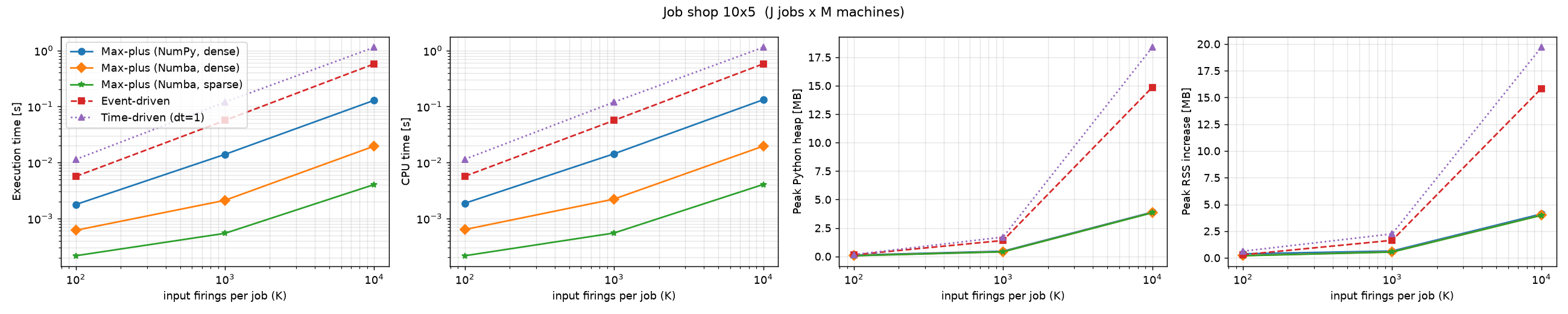}
\caption{Benchmark on a $10\times5$ job shop, as a function of the number of input firings per job $K$ (logarithmic axes). From left to right: execution time, CPU time, peak Python heap, and peak RSS increase. The two rows are two independent benchmark runs, used respectively as the event-driven and as the time-driven reference of the speedups.}
\label{fig:bench}
\end{figure}

We simulate repetitive job shops of four sizes ($J\times M$): $3\times3$, $6\times4$, $10\times5$ and $20\times10$, with $K\in\{10^2,10^3,10^4\}$ input firings per job (that is, $J\cdot M\cdot K$ operation completions). Five simulators are compared on the same instance and the same machine orders:
\begin{itemize}[leftmargin=*]
\item \emph{Event-driven}: the reference discrete-event simulator, with an event list.
\item \emph{Time-driven} ($dt=1$): a fixed-step simulator that advances the clock one unit at a time.
\item \emph{Max-plus, NumPy, dense}; \emph{max-plus, Numba, dense}; \emph{max-plus, Numba, sparse}: the three implementations of the recursion of Section~\ref{sec:model}.
\end{itemize}
We record wall-clock time, CPU time, peak Python heap and peak resident-set-size increase. The speedup is the wall time of the reference simulator divided by that of the max-plus simulator. Numba compilation (about $0.13$--$0.15$\,s, once per process) is measured separately and not included in the wall time; Table~\ref{tab:benchbet} (Appendix~\ref{app:bench}) gives the number of simulations after which it is recovered. Two independent runs of the benchmark were made, one normalised to the event-driven and one to the time-driven simulator (Figure~\ref{fig:bench}). A run is accepted (flag ``ok'') only if every simulator returns the same makespan; this was the case for all rows. In this benchmark all durations and input dates are integers, so the time-driven simulator with $dt=1$ is exact; with real-valued dates it would round every start up to the next tick and simulate a slightly different system.

\begin{table}[t]
\centering
\caption{Simulation benchmark at $K=10^4$ input firings per job (job shops, $J$ jobs $\times$ $M$ machines). Wall-clock time of the two reference simulators and of the Numba sparse max-plus simulator, and speedup of each max-plus implementation over the event-driven (``vs ev.'') and over the time-driven (``vs td.'') simulator. The ``vs ev.'' speedups and the event-driven and sparse times come from the first benchmark run; the ``vs td.'' speedups and the time-driven times from the second run, in which the max-plus simulators were timed again, so the ``vs td.'' column cannot be recomputed exactly from the times shown. JIT compilation ($\approx0.14$\,s, once) is excluded (see text). All simulators return the same makespan. Results for $K=10^2,10^3$ are in Appendix~\ref{app:bench}.}
\label{tab:bench}
\scriptsize
\setlength{\tabcolsep}{4pt}
\begin{tabular}{l rrr rr rr rr}
\toprule
 & \multicolumn{3}{c}{Wall-clock time} & \multicolumn{2}{c}{NumPy, dense} & \multicolumn{2}{c}{Numba, dense} & \multicolumn{2}{c}{Numba, sparse}\\
\cmidrule(lr){2-4}\cmidrule(lr){5-6}\cmidrule(lr){7-8}\cmidrule(lr){9-10}
Shop & event & time-dr. & sparse & vs ev. & vs td. & vs ev. & vs td. & vs ev. & vs td.\\
\midrule
$3\times3$ & 106.19 ms & 190.12 ms & 0.88 ms & 1.9 & 3.5 & 108 & 196 & 120 & 198 \\
$6\times4$ & 290.97 ms & 418.39 ms & 2.02 ms & 3.5 & 5.0 & 62 & 89 & 144 & 207 \\
$10\times5$ & 613.00 ms & 1.14 s & 4.12 ms & 4.5 & 8.8 & 31 & 58 & 149 & 282 \\
$20\times10$ & 2.40 s & 5.00 s & 12.36 ms & 3.8 & 8.6 & 6.2 & 14 & 194 & 419 \\
\bottomrule
\end{tabular}
\end{table}

\subsection{Results}
Table~\ref{tab:bench} and Figure~\ref{fig:bench} give the main results; all numbers are in Appendix~\ref{app:bench}.
\begin{itemize}[leftmargin=*]
\item \emph{Speed.} At $K=10^4$ the Numba sparse simulator is $120\times$ to $194\times$ faster than the event-driven simulator and $198\times$ to $419\times$ faster than the time-driven one, over the four sizes. The maximum, $194\times$ and $419\times$, is reached for the largest shop, $20\times10$. For the $10\times5$ shop of Figure~\ref{fig:bench} the factors are $149\times$ and $282\times$ (execution time of $4.1$\,ms against $0.61$\,s and $1.1$\,s). In throughput, the event-driven simulator completes about $8\times10^5$ operations per second whatever the size, and the sparse max-plus simulator between $1.0\times10^8$ and $1.6\times10^8$ at $K=10^4$.
\item \emph{The factor grows with $K$ and with the size, for the sparse variant.} For the sparse implementation the speedup over the event-driven simulator rises from $8\times$ to $35\times$ at $K=10^2$ to $120\times$--$194\times$ at $K=10^4$, because fixed costs (call overhead, array allocation) are amortised, and it increases with the shop size at $K=10^4$ ($120\times$, $144\times$, $149\times$, $194\times$). The time-driven simulator is slower than the event-driven one by a factor of $1.4$ to $2.1$, since it performs one step per time unit; this factor depends on the time scale (durations ten times larger would make it ten times slower), so the speedups over the time-driven simulator are illustrative rather than intrinsic.
\item \emph{Compilation and sparsity both matter.} The NumPy dense implementation gains only $1.9\times$ to $4.5\times$ over the event-driven simulator and $3.5\times$ to $8.8\times$ over the time-driven one at $K=10^4$. Compiling the dense recursion with Numba gives $108\times$ ($3\times3$) but this falls to $31\times$ ($10\times5$) and $6.2\times$ ($20\times10$), as the dense matrices grow with the square of the number of operations (Remark~\ref{rem:cost}); only the sparse variant keeps and increases its advantage. The max-plus structure alone, without compilation, is therefore \emph{not} what gives two orders of magnitude; it is what makes compilation and sparse storage possible.
\item \emph{Memory.} At $K=10^4$ the peak Python heap of the Numba sparse simulator is $3.8$\,MB for $10\times5$ and $15$\,MB for $20\times10$, against $14.9$\,MB and $39.9$\,MB (event-driven) and $18.4$\,MB and $47.0$\,MB (time-driven); over the four sizes it is 2.6 to 4.6 times lower than that of the event-driven simulator and 3.1 to 6.0 times lower than that of the time-driven one. For short runs the ordering can reverse: at $K=10^2$ on the $20\times10$ shop the dense max-plus variants use more heap than the event-driven simulator. Peak Python heap is measured with \texttt{tracemalloc}, which may not see all allocations made inside compiled code; the peak RSS increase, which does, shows the same ordering (Figure~\ref{fig:bench}).
\item \emph{Compilation cost.} The one-off JIT time ($\approx0.14$\,s) exceeds the time of a single short simulation, but it is recovered after a single run for $K=10^4$ in the shops of $6\times4$ and larger, and after $7$ to $159$ runs for $K=10^2$ (Table~\ref{tab:benchbet}). In RL, where millions of simulations are run in one process, it is negligible.
\end{itemize}

\paragraph{What this benchmark does not show.} (i) The reference simulators are our own implementations; we did not compare with a compiled event-driven simulator or with an established simulation library, which would reduce the gap; part of the speedup reflects compiled versus interpreted code, which is why the NumPy row is given. (ii) Only job shops (one family of NCTEGs) were benchmarked. (iii) The benchmark measures the simulator over long horizons, not an RL training loop, in which the simulator is called one round at a time and policy evaluation and the gradient update come in addition; Section~\ref{sec:cost} shows that in our application the simulator takes only 15--27\% of the episode time, so the end-to-end gain is much smaller than the simulation speedup. (iv) Single process, single thread; we did not test parallel or GPU execution, which the independence of rollouts would allow. (v) Equality of makespans is a necessary but not a complete check of equivalence of the simulators.

\section{Application: Reinforcement Learning for Job Shop Scheduling}\label{sec:app}

Repetitive job shops are an NCTEG in which each machine is a conflict place, and the conflict resolution is the order in which the operations use the machine \cite{addad2012}. This section applies the framework of Section~\ref{sec:model} to the problem of choosing release dates and machine orders so that completion dates track due dates. It illustrates the approach on one concrete problem and does not aim at a state-of-the-art comparison with the job-shop literature.

\subsection{Problem Formulation}\label{sec:problem}

\subsubsection{Repetitive job shop as an NCTEG}
A job shop has $J$ jobs and $M$ machines. Job $j$ visits every machine once, in its own route $\mu(j,1),\dots,\mu(j,M)$, with nominal durations $p_{jk}$. The jobs are processed repeatedly: round $n=0,\dots,N-1$ processes one batch of every job. Each machine is a conflict place; the order in which the operations of round $n$ use it is the conflict resolution $d_n$. Let $e_{jk}(n+1)$ be the completion date of operation $k$ of job $j$ in round $(n+1)$. For a fixed set of machine orders,
\[
e_{jk}(n+1)=p_{jk}+\max\Big(\underbrace{e_{j,k-1}(n+1)}_{\text{job precedence}},\ \underbrace{e_{\pi(jk)}(n+1)\ \text{or}\ \max_{o\in\mathcal{O}_m}e_{o}(n)}_{\text{machine}},\ \underbrace{e_{j,M}(n)\vee u_j(n+1)}_{k=1}\Big),
\]
where $\pi(jk)$ is the predecessor of the operation on its machine in the chosen order (or, if it is the first one in the round, the machine becomes free when the last operation of the previous round, among the set $\mathcal{O}_m$ of operations on that machine, completes); the $k=1$ term states that a job restarts after its previous batch has completed and after its release date $u_j(n+1)$. This is exactly \eqref{eq:smpl} with state $x=e$, matrix $A_0(d)$ holding the same-round arcs, $A_1$ the arcs from the previous round (independent of $d$, because the machine is free after the \emph{latest} completion on it), $B$ the release dates, and $y_j=e_{jM}$, the completion date of job $j$.

\subsubsection{Control objective and MDP}
Given due dates $Z(n)$, the aim is to choose release dates $u(n)$ and machine orders $d_n$ minimizing
\begin{equation}
\mathcal{J}=\sum_{n=0}^{N-1}\sum_{j=1}^{J}w(\delta_j(n))\,\delta_j(n)^2,\qquad \delta_j(n)=y_j(n)-Z_j(n),
\label{eq:cost}
\end{equation}
with $w=w_{\mathrm{late}}=3$ if $\delta>0$ and $w=w_{\mathrm{early}}=1$ otherwise (just-in-time cost). The MDP has state $s_n=(x(n)-r_n,\,Z(n)-r_n)$ with reference date $r_n$ (Proposition~\ref{prop:equiv}), action $(u(n),d_n)$, deterministic max-plus transition (stochastic when durations are random), and reward $-\sum_j w(\delta_j)\delta_j^2$.

\subsection{Method}\label{sec:method}

\paragraph{Conflict resolution by sequential decisions.} The set of machine orders is $(J!)^M$ and many of them are deadlocked: opposite orders on two machines can create a token-free circuit in $A_0$. We therefore build $d_n$ sequentially with a Giffler--Thompson scheme. At every step, among the next unscheduled operations of all jobs, the one with the earliest possible completion date identifies a machine $m^\star$, which is the conflict place being resolved. The \emph{conflict set} is the set of operations on $m^\star$ that could start before that completion date. The policy chooses one operation of the conflict set, which is appended to the order of $m^\star$. When the conflict set is a singleton, no decision is taken.

\begin{proposition}[Liveness]\label{prop:live}
Let the machine orders be the restrictions, to each machine, of one sequence of all operations of the round in which the operations of every job appear in route order. Then $A_0(d)$ has no circuit, hence it is nilpotent and $A_0(d)^*$ exists. In particular, every set of machine orders produced by the scheme above is live.
\end{proposition}
\begin{proof}
Both the job-precedence and the machine-order arcs of $A_0$ go from an earlier to a later element of the sequence, so $A_0$ has no circuit and is nilpotent. The scheme outputs such a sequence: each job's operations are appended in route order, and the order of every machine is the subsequence of its operations.
\end{proof}
Liveness therefore comes from encoding a mode by a sequence of operations, not from the Giffler--Thompson rule itself; the random job sequences used by one of the baselines below are also live. What the scheme adds is that the schedules are \emph{active} for the nominal durations and that each decision is a choice among a few operations. For a regular objective an optimal schedule can be found among active ones, but the cost~\eqref{eq:cost} is not regular, since it penalizes earliness, and restricting to active schedules may exclude the optimum; here idle time can only be inserted through the release dates.

\paragraph{Policy.} For an operation $o=(j,k)$ in a conflict set, the features $\varphi(o)\in\mathbb{R}^7$ are, relative to the reference date and scaled by the period $T$: earliest start; projected slack $Z_j-(\text{start}+p_{jk}+\text{remaining work})$; duration; remaining work of the job after the operation; due date; the negative part of the slack; and the fraction of remaining operations. The probability of choosing $o$ is $\mathrm{softmax}_o(\theta^\top\varphi(o))$. Most classical dispatching rules (earliest start/FIFO, earliest due date, least slack, shortest processing time, most work remaining) are particular weight vectors $\theta$ of this same family, so these baselines lie inside the policy class. The continuous part sets the release offset: $u_j(n)=\max\big(u_j(n-1),\,Z_j(n)-P_j-\tau\big)$ with $P_j=\sum_kp_{jk}$ and $\tau=0.5\,T\,\mathrm{softplus}(\rho)$, $\rho\sim\mathcal{N}(b,\sigma^2)$ per job and round, which keeps release dates non-decreasing by construction.

\paragraph{Learning.} We use REINFORCE \cite{williams1992} with reward-to-go. Each iteration draws $n_{\mathrm{inst}}$ fresh random job shops from the instance distribution and, for each, $R$ rollouts sharing the same noise realization; the baseline is the mean return over the rollouts of the same instance at the same round (so that differences are attributable to the decisions), and advantages are divided by their standard deviation. The score-function gradient of a conflict decision is $\varphi(o)-\sum_{o'}\pi(o')\varphi(o')$, and that of the release offset is $(\rho-b)/\sigma^2$. Updates use Adam \cite{kingma2015}; $\sigma$ decays from $0.5$ to $0.1$. At test time the policy is deterministic (argmax over the conflict set, mean release offset).

\paragraph{Decision-time search (RL + search).} Because a max-plus step is cheap, the learned policy can also be combined with the model when it is deployed. At every round, the greedy schedule and $K$ schedules sampled from the policy (softmax at temperature $\tau_s$) are evaluated with the max-plus recursion on the nominal durations, and the schedule applied is the one that minimizes the cost of the round plus a terminal term $\lambda\,T\sum_m\big(b_m-\bar Z(n)\big)$, where $b_m$ is the date at which machine $m$ becomes free at the end of the round and $\bar Z(n)$ the mean due date of the round. The terminal term prevents the choice from trading the current round against the next one, which the learned policy accounts for but a one-round criterion does not. Release dates are those of the greedy policy. The pair $(\lambda,\tau_s)\in\{0,1,3\}\times\{1,3\}$ is tuned for each training seed on the tuning instances; $K\in\{1,50,100\}$.

\paragraph{Implementation.} The round simulator and the Giffler--Thompson decoding with the linear policy are compiled with Numba \cite{lam2015}; the NumPy implementation is kept as reference. Exploration in the compiled decoder consumes the random stream exactly as the NumPy code does, so the two implementations produce the same trajectories; a check at start-up verifies schedules, gradients and random state on random cases, and the experiments were run with both (Section~\ref{sec:cost}). Before running, the code also checks that the max-plus matrix model~\eqref{eq:smpl} (with explicit Kleene star) and the vectorized evaluator agree on random modes.

\subsection{Experiments}\label{sec:exp}

\subsubsection{Setup}

\paragraph{Instances.} Routes are uniformly random permutations of the machines, durations are uniform integers in $\{1,\dots,9\}$, the period is $T=1.1\max_m(\text{load of machine }m)$, and the due dates are $Z_j(n)=nT+P_j+U_j(n)$ with $U_j(n)\sim\mathcal{U}(0,T)$, over $N=8$ rounds. We consider six settings: $3\times3$, $6\times4$ and $10\times5$ shops (jobs $\times$ machines), each with exact durations and with noisy durations, where the realized durations are the nominal ones multiplied by $\mathcal{U}(0.7,1.3)$ in every operation and round. All controllers decide each round using nominal durations; the realized durations are revealed after the round.

\paragraph{Baselines.} (i) \emph{Random order}: random choice in each conflict set. (ii)--(vi) Five dispatching rules applied to the conflict sets: earliest start time (EST/FIFO), earliest due date (EDD), least slack, shortest processing time (SPT), and most work remaining (MWKR, remaining work including the current operation). We also report the variant of MWKR that counts only the work after the current operation, used in an earlier version of this work. (vii) \emph{MPC-1}: a one-round-lookahead controller on the nominal model that searches job sequences and picks the one with minimal cost for the current round: exhaustive search over all job sequences for $3\times3$ shops (1\,680 sequences) and the best of 3\,000 uniformly random sequences for $6\times4$ and $10\times5$ shops. (viii) \emph{MPC-1+}: a non-myopic one-round MPC whose candidate pool contains the schedules of the dispatching rules and 300 Giffler--Thompson schedules with random priorities (active schedules), and whose objective adds to the round cost the terminal term $\lambda\,T\sum_m(b_m-\bar Z(n))$ of the decision-time search. MPC-1 is myopic: the earliness penalty rewards delaying jobs that would finish early, which keeps the machines busy into the next round, and uniformly random sequences are mostly non-active; MPC-1+ removes both weaknesses. (ix) \emph{CEM}: the cross-entropy method over $(\theta,b)$ of the same linear policy, with greedy evaluation (population 24, 6 elite, 6 fresh instances per iteration, fitness the mean log cost), using the same number of training episodes as REINFORCE; it tests whether the gradient method matters or only the policy class. Every baseline except CEM gets its release offset, a constant fraction of $T$ in $\{0,0.1,0.25,0.5,0.75,1,1.25,1.5,1.75,2\}$, tuned on a separate set of 12 tuning instances; for MPC-1+ the offset and $\lambda\in\{0,0.3,1,3\}$ are tuned jointly. The RL policy learns its offset.

\paragraph{Ablation: same release offset.} The policy learns its release offset by gradient while the baselines choose theirs from a grid, so part of a difference could come from the continuous part of the policy and not from routing. To isolate routing, we re-evaluate every baseline on the same test instances and the same noise realizations with the release offset learned by the policy (as a fraction of $T$). For each training seed, each baseline is run with the offset learned by that seed, and the reported baseline cost is the mean over seeds; the RL cost is the one of the main tables. The per-seed reduction compares the RL policy of a seed with the best baseline run with the offset of that same seed. The baselines use no other component of the policy.

\paragraph{Protocol.} Evaluation is performed on unseen instances: 40 test instances per setting, with the same noise for all methods; the settings with exact durations are evaluated once per instance and those with noise on 5 noise realizations per instance. The RL policy is trained from $\theta=0$ (uniformly random choices) with 5 seeds; each training uses 300 iterations, with $8\times8$ rollouts per iteration (8 fresh instances, 8 rollouts each) for the $3\times3$ shops and $6\times6$ rollouts for the $6\times4$ and $10\times5$ shops; learning rate $0.05$. The hyperparameters were not tuned on the test instances. Reported RL numbers are averages over seeds. Cost tables give the mean over test instances with a 95\% normal confidence interval. For paired comparisons we report the relative reduction of the mean cost, the 95\% percentile bootstrap interval of the mean paired difference over test instances (10\,000 resamples), the share of test instances on which a method has the lower cost (``win''), and the median of the per-instance cost ratio; a reduction is called significant when the bootstrap interval excludes zero. Example instances and schedules are drawn in Appendix~\ref{app:examples}.

\subsubsection{Results}

\begin{figure}[t]
\centering
\includegraphics[width=\textwidth]{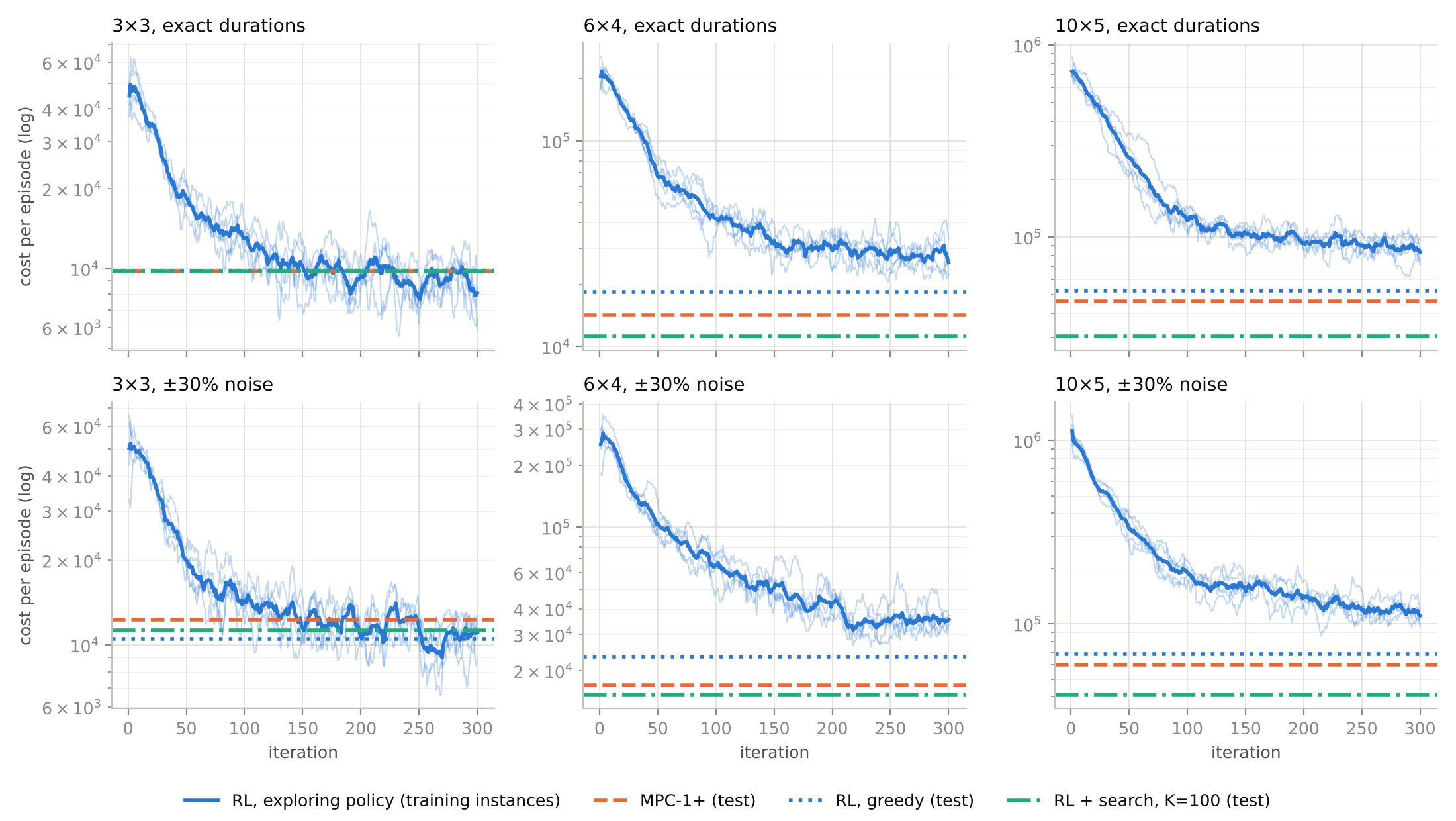}
\caption{Training curves of the RL policy for the six settings: cost per episode of the exploring policy on the fresh random instances of each iteration (moving average over 10 iterations; thin lines are the 5 seeds, the thick line is their mean), with the test cost of MPC-1+, of the greedy RL policy and of RL + search ($K=100$) as horizontal lines. Training and test instances differ and the training policy explores, so the curves and the horizontal lines are only indicative. Logarithmic vertical axis. For $3\times3$ with exact durations the three horizontal lines nearly coincide. The original curves and bar charts of each setting are in Appendix~\ref{app:allfigs}.}
\label{fig:curves}
\end{figure}

Tables~\ref{tab:exact} and~\ref{tab:noisy} give the mean cost~\eqref{eq:cost} per episode (sum over 8 rounds) for exact and noisy durations; the mean total tardiness is in Appendix~\ref{app:tard}. Table~\ref{tab:classic} compares the policy with the classical baselines and with CEM, Table~\ref{tab:mpcp} with MPC-1+, and Figure~\ref{fig:curves} shows the training curves.

\begin{table}[H]
\centering
\caption{Mean cost per episode ($\pm$ 95\% CI over the 40 test instances), exact durations; lower is better. RL, RL + search and CEM are averages over 5 training seeds. Bold: lowest mean cost; underlined: best baseline when it is not the lowest. $^{a}$~Work after the current operation only. $^{b}$~Exhaustive over 1\,680 job sequences for $3\times3$; best of 3\,000 random sequences for $6\times4$ and $10\times5$. $^{c}$~Rule schedules and 300 random-priority Giffler--Thompson schedules, terminal cost. $^{d}$~Cross-entropy search over the weights of the linear policy.}
\label{tab:exact}
\scriptsize
\setlength{\tabcolsep}{4pt}
\begin{tabular}{l rrr}
\toprule
Method & $3\times3$ & $6\times4$ & $10\times5$\\
\midrule
Random order & 34\,321 $\pm$ 11\,855 & 197\,753 $\pm$ 52\,924 & 419\,401 $\pm$ 90\,155 \\
EST / FIFO & 12\,910 $\pm$ 8\,632 & 68\,874 $\pm$ 33\,751 & 197\,822 $\pm$ 54\,862 \\
EDD & 45\,616 $\pm$ 18\,265 & 310\,314 $\pm$ 88\,266 & 782\,608 $\pm$ 146\,610 \\
Least slack & 42\,644 $\pm$ 21\,002 & 206\,357 $\pm$ 50\,691 & 514\,502 $\pm$ 111\,598 \\
SPT & 68\,049 $\pm$ 31\,927 & 467\,527 $\pm$ 150\,513 & 1\,483\,451 $\pm$ 364\,908 \\
Most work remaining (MWKR) & 16\,720 $\pm$ 10\,590 & 40\,303 $\pm$ 10\,444 & 159\,671 $\pm$ 44\,506 \\
MWKR, after operation$^{a}$ & 20\,143 $\pm$ 11\,616 & 48\,755 $\pm$ 18\,183 & 122\,120 $\pm$ 52\,364 \\
MPC-1$^{b}$ & 18\,039 $\pm$ 8\,424 & 40\,976 $\pm$ 16\,728 & 210\,225 $\pm$ 45\,844 \\
MPC-1+$^{c}$ & \textbf{9\,728} $\pm$ 6\,569 & \underline{14\,250} $\pm$ 7\,934 & \underline{46\,444} $\pm$ 14\,404 \\
CEM, same policy class$^{d}$ & 12\,511 $\pm$ 6\,539 & 24\,775 $\pm$ 7\,931 & 64\,046 $\pm$ 16\,607 \\
\midrule
\textbf{RL (ours)} & 9\,807 $\pm$ 5\,226 & 18\,428 $\pm$ 3\,235 & 52\,433 $\pm$ 8\,116 \\
RL + search, $K=1$ & 9\,846 $\pm$ 5\,407 & 16\,875 $\pm$ 3\,340 & 49\,883 $\pm$ 6\,970 \\
RL + search, $K=50$ & 9\,809 $\pm$ 6\,338 & 12\,725 $\pm$ 5\,807 & 34\,311 $\pm$ 4\,419 \\
RL + search, $K=100$ & 9\,777 $\pm$ 6\,350 & \textbf{11\,251} $\pm$ 3\,973 & \textbf{30\,489} $\pm$ 2\,991 \\
\bottomrule
\end{tabular}
\end{table}

\begin{table}[H]
\centering
\caption{Mean cost per episode ($\pm$ 95\% CI over test instances), durations with $\pm30\%$ noise; same layout and notes as Table~\ref{tab:exact}.}
\label{tab:noisy}
\scriptsize
\setlength{\tabcolsep}{4pt}
\begin{tabular}{l rrr}
\toprule
Method & $3\times3$ & $6\times4$ & $10\times5$\\
\midrule
Random order & 37\,324 $\pm$ 11\,677 & 230\,688 $\pm$ 49\,006 & 652\,430 $\pm$ 113\,504 \\
EST / FIFO & 13\,827 $\pm$ 8\,500 & 72\,102 $\pm$ 29\,007 & 264\,044 $\pm$ 78\,852 \\
EDD & 52\,689 $\pm$ 18\,910 & 392\,341 $\pm$ 98\,404 & 1\,071\,872 $\pm$ 166\,556 \\
Least slack & 46\,290 $\pm$ 20\,222 & 242\,041 $\pm$ 62\,363 & 749\,984 $\pm$ 125\,300 \\
SPT & 73\,368 $\pm$ 31\,429 & 520\,913 $\pm$ 158\,283 & 1\,771\,868 $\pm$ 408\,315 \\
Most work remaining (MWKR) & 22\,080 $\pm$ 13\,053 & 60\,505 $\pm$ 15\,395 & 202\,012 $\pm$ 55\,234 \\
MWKR, after operation$^{a}$ & 22\,078 $\pm$ 12\,479 & 62\,487 $\pm$ 21\,766 & 142\,485 $\pm$ 57\,994 \\
MPC-1$^{b}$ & 21\,933 $\pm$ 9\,580 & 62\,684 $\pm$ 20\,256 & 269\,912 $\pm$ 46\,458 \\
MPC-1+$^{c}$ & \underline{12\,264} $\pm$ 8\,188 & \underline{17\,015} $\pm$ 5\,798 & \underline{60\,058} $\pm$ 8\,595 \\
CEM, same policy class$^{d}$ & 14\,104 $\pm$ 7\,426 & 28\,997 $\pm$ 8\,135 & 84\,098 $\pm$ 21\,473 \\
\midrule
\textbf{RL (ours)} & \textbf{10\,486} $\pm$ 5\,737 & 23\,304 $\pm$ 4\,371 & 68\,226 $\pm$ 8\,077 \\
RL + search, $K=1$ & 10\,954 $\pm$ 6\,551 & 21\,389 $\pm$ 4\,298 & 67\,561 $\pm$ 8\,351 \\
RL + search, $K=50$ & 11\,185 $\pm$ 7\,071 & 15\,679 $\pm$ 5\,425 & 43\,234 $\pm$ 4\,333 \\
RL + search, $K=100$ & 11\,246 $\pm$ 7\,265 & \textbf{15\,245} $\pm$ 5\,531 & \textbf{41\,053} $\pm$ 4\,097 \\
\bottomrule
\end{tabular}
\end{table}

\begin{table}[H]
\centering
\caption{RL (greedy policy) versus the best classical baseline (dispatching rules, random order and MPC-1) and versus CEM in the same policy class. Reduction = relative reduction of the mean cost (positive favours RL); bootstrap interval of the mean paired difference (baseline minus RL); ``win'' = share of the 40 test instances where RL has the lower cost.}
\label{tab:classic}
\scriptsize
\setlength{\tabcolsep}{3.5pt}
\begin{tabular}{l l r r r r r}
\toprule
 & \multicolumn{4}{c}{Best classical baseline} & \multicolumn{2}{c}{CEM}\\
\cmidrule(lr){2-5}\cmidrule(lr){6-7}
Setting & Baseline & Reduction & Bootstrap 95\% CI & Win & Reduction & Win\\
\midrule
$3\times3$, exact & EST/FIFO & 24.0\% & [$-$873,\ 8\,516] & 32\% & 21.6\% & 60\% \\
$3\times3$, $\pm30\%$ & EST/FIFO & 24.2\% & [335,\ 7\,786] & 68\% & 25.6\% & 62\% \\
$6\times4$, exact & MWKR & 54.3\% & [12\,335,\ 32\,666] & 78\% & 25.6\% & 60\% \\
$6\times4$, $\pm30\%$ & MWKR & 61.5\% & [23\,008,\ 51\,947] & 80\% & 19.6\% & 57\% \\
$10\times5$, exact & MWKR, after op. & 57.1\% & [33\,051,\ 126\,516] & 90\% & 18.1\% & 48\% \\
$10\times5$, $\pm30\%$ & MWKR, after op. & 52.1\% & [30\,641,\ 136\,596] & 75\% & 18.9\% & 55\% \\
\bottomrule
\end{tabular}
\end{table}

\begin{table}[H]
\centering
\caption{RL and RL + search versus MPC-1+, the best baseline in every setting. Relative reduction of the mean cost (positive favours the learned controller) and share of test instances where it has the lower cost. $^{*}$: the bootstrap 95\% interval of the mean paired difference excludes zero. Intervals and median cost ratios are in Table~\ref{tab:mpcp_detail} (Appendix~\ref{app:param}).}
\label{tab:mpcp}
\scriptsize
\setlength{\tabcolsep}{4pt}
\begin{tabular}{l rr rr rr}
\toprule
 & \multicolumn{2}{c}{RL (greedy)} & \multicolumn{2}{c}{RL + search, $K=50$} & \multicolumn{2}{c}{RL + search, $K=100$}\\
\cmidrule(lr){2-3}\cmidrule(lr){4-5}\cmidrule(lr){6-7}
Setting & Reduction & Win & Reduction & Win & Reduction & Win\\
\midrule
$3\times3$, exact & $-$0.8\% & 50\% & $-$0.8\% & 65\% & $-$0.5\% & 62\% \\
$3\times3$, $\pm30\%$ & +14.5\% & 38\% & +8.8\%$^{*}$ & 57\% & +8.3\%$^{*}$ & 55\% \\
$6\times4$, exact & $-$29.3\% & 15\% & +10.7\% & 35\% & +21.0\% & 38\% \\
$6\times4$, $\pm30\%$ & $-$37.0\%$^{*}$ & 10\% & +7.9\% & 65\% & +10.4\%$^{*}$ & 75\% \\
$10\times5$, exact & $-$12.9\% & 25\% & +26.1\%$^{*}$ & 35\% & +34.4\%$^{*}$ & 38\% \\
$10\times5$, $\pm30\%$ & $-$13.6\%$^{*}$ & 30\% & +28.0\%$^{*}$ & 90\% & +31.6\%$^{*}$ & 90\% \\
\bottomrule
\end{tabular}
\end{table}

\paragraph{Observations.}
\begin{itemize}[leftmargin=*]
\item \emph{Against the classical baselines and CEM.} The learned policy lowers the mean cost by 24.0\% and 24.2\% ($3\times3$), 54.3\% and 61.5\% ($6\times4$) and 57.1\% and 52.1\% ($10\times5$) relative to the best of the dispatching rules, random order and MPC-1 (exact and noisy durations, respectively; Table~\ref{tab:classic}). The bootstrap interval excludes zero in five settings; in the $3\times3$ shops with exact durations it does not, and the policy beats EST/FIFO on only 32\% of the instances, so the lower mean comes from large gains on a minority of instances where EST/FIFO performs badly. The test cost of every single seed is below that of the best classical baseline of its setting, by at least 21\%. Against CEM, which searches the same seven weights and offset with the same number of episodes, the policy is 18--26\% better, significantly in the $3\times3$ and $6\times4$ settings; REINFORCE is therefore a better use of the simulation budget than derivative-free search in this policy class.
\item \emph{MPC-1+ is by far the strongest baseline.} It is the best baseline in all six settings, 44--78\% below MPC-1. Its two changes (candidate pool of active schedules, terminal cost) matter more than any other choice among the baselines; MPC-1 is worse than EST/FIFO on the $3\times3$ shops although it searches all job sequences, a symptom of its myopia. Against MPC-1+ the greedy policy is equal on the $3\times3$ shops with exact durations ($-0.8\%$, 50\% of the instances), better on average with noise ($+14.5\%$, not significant, 38\% of the instances), and worse on the $6\times4$ and $10\times5$ shops, by 12.9\% to 37.0\%, significantly in the two noisy settings (Table~\ref{tab:mpcp}).
\item \emph{But it decides much faster.} The policy takes 0.05\,ms per round against 7.4 to 9.1\,ms for MPC-1+ (compiled implementations, Table~\ref{tab:compute}), 150 to 190 times less.
\item \emph{Combining the two is best.} RL + search with $K=100$ has a lower mean cost than MPC-1+ in five settings, by 8.3\% ($3\times3$, noisy), 21.0\% and 10.4\% ($6\times4$) and 34.4\% and 31.6\% ($10\times5$), significantly in four of them (all but $6\times4$ with exact durations); it is equal on $3\times3$ with exact durations ($-0.5\%$). The gain is broad for $6\times4$ and $10\times5$ with noise (75\% and 90\% of the instances) but not for $10\times5$ with exact durations, where RL + search wins on only 38\% of the instances and its median cost ratio to MPC-1+ is 1.26: the lower mean comes from instances where MPC-1+ performs badly. RL + search takes 2.1 to 3.0\,ms per round, about a third of the time of MPC-1+, although both are one-round controllers on the same model: the candidates proposed by the learned policy are better than the random-priority candidates of MPC-1+, so fewer are needed.
\item \emph{Search helps on larger shops, not on the smallest.} Relative to the greedy policy, the search lowers the mean cost by 31--42\% for $K=50$ and $K=100$ on the $6\times4$ and $10\times5$ shops (winning on 92--100\% of the instances), and already by 1--8\% with a single sampled schedule ($K=1$). On the $3\times3$ shops it does not help ($+0.3\%$ with exact durations, $-7.2\%$ with noise, neither significant). With few conflicts per round, the greedy policy is close to what a one-round criterion can find; we did not investigate this further.
\item \emph{Seeds.} The standard deviation of the 5 seed means is 2.5--2.6\% of the mean for the $3\times3$ shops, 5.9--6.9\% for the $6\times4$ shops and for $10\times5$ with noise, and 15.3\% for $10\times5$ with exact durations, where two of the five seeds are markedly worse (Appendix~\ref{app:seeds}).
\item \emph{Training length.} The cost of the exploring policy on the training instances still decreases by up to 12\% over the last 50 iterations in five of the six settings (Figure~\ref{fig:curves}), but the cost of the greedy policy on the tuning instances no longer improves between iterations 250 and 300 (changes between $-0.4\%$ and $+8.7\%$). Longer training of the same linear policy is therefore unlikely, by itself, to close the gap with MPC-1+ on the larger shops.
\item \emph{The learned weights are interpretable and consistent.} Across seeds and settings (Table~\ref{tab:weights}, Appendix~\ref{app:seeds}) the policy combines a strong preference for operations that can start early (weight $-9.4$ to $-11.5$ on the earliest start, FIFO-like) with a preference for jobs with much remaining work and many remaining operations (weights $+5.4$ to $+6.0$ and $+4.9$ to $+6.0$, MWKR-like) and for short operations ($-1.1$ to $-2.9$, SPT-like). The standard deviation over seeds of every weight is at most 0.8. No single rule provides this mixture. The learned release offsets are between $0.94\,T$ and $1.30\,T$ on average, with a seed-to-seed range of $0.66\,T$ to $1.84\,T$ (Table~\ref{tab:offsets}).
\item \emph{Noise.} The relative position of the methods is not systematically different with and without noise. All controllers decide routing at the beginning of each round from nominal durations and react to noise only through the state of the next round, so the experiments do not show an advantage specific to uncertainty.
\item \emph{Due-date rules.} Due-date oriented rules are poor in this setting: EDD and least slack are worse than FIFO by a factor 3--4 in the $3\times3$ shops, and EDD is 6 to 8 times worse than the best rule in the larger shops, plausibly because in this repetitive setting keeping the machines busy matters more than due-date urgency and the quadratic cost penalizes the resulting backlog; we did not test this explanation.
\end{itemize}

\subsubsection{Ablation: the effect of routing alone}\label{sec:abl}

The comparison above leaves one confounder: the policy learns its release offset whereas the baselines pick theirs from a grid. Table~\ref{tab:abl} gives the result of the ablation described in the setup, in which every baseline uses the release offset learned by the policy, so that remaining differences are attributable to the conflict resolutions. The costs of all baselines are in Appendix~\ref{app:abl}.

\begin{table}[t]
\centering
\caption{Ablation with the same release offset: RL versus MPC-1+, which remains the best baseline in every setting when every baseline uses the release offset learned by the policy. ``Own offset'' and ``learned offset'' are the mean costs of MPC-1+ with its grid-tuned offset (as in Tables~\ref{tab:exact}--\ref{tab:noisy}) and with the learned offset (mean over the 5 seeds); the reduction (positive favours RL) is computed against the latter; ``win'' is the share of the 40 test instances where RL has the lower cost; the last column gives the range over seeds of the per-seed reduction.}
\label{tab:abl}
\scriptsize
\setlength{\tabcolsep}{3pt}
\begin{tabular}{l r r r r r r}
\toprule
Setting & Own offset & Learned offset & RL & Reduction & Win & Per-seed reduction\\
\midrule
$3\times3$, exact & 9\,728 & 9\,625 & 9\,807 & $-$1.9\% & 38\% & $-$4.0\% to $-$0.6\% \\
$3\times3$, $\pm30\%$ & 12\,264 & 12\,260 & 10\,486 & +14.5\% & 35\% & +11.0\% to +16.9\% \\
$6\times4$, exact & 14\,250 & 13\,366 & 18\,428 & $-$37.9\% & 10\% & $-$55.9\% to $-$20.8\% \\
$6\times4$, $\pm30\%$ & 17\,015 & 19\,497 & 23\,304 & $-$19.5\% & 35\% & $-$24.5\% to $-$13.4\% \\
$10\times5$, exact & 46\,444 & 44\,547 & 52\,433 & $-$17.7\% & 22\% & $-$48.2\% to +14.6\% \\
$10\times5$, $\pm30\%$ & 60\,058 & 57\,255 & 68\,226 & $-$19.2\% & 15\% & $-$29.6\% to $-$13.5\% \\
\bottomrule
\end{tabular}
\end{table}

\paragraph{Observations.}
\begin{itemize}[leftmargin=*]
\item \emph{The release offset does not change the conclusions.} Replacing the grid-tuned offset of a baseline by the learned one changes its cost by between $-17.1\%$ and $+14.6\%$ (Appendix~\ref{app:abl}), and MPC-1+ remains the best baseline in all six settings. Against it, the policy is $1.9\%$ worse ($3\times3$, exact), 14.5\% better ($3\times3$, noisy) and 17.7\% to 37.9\% worse on the larger shops, close to the main comparison. The gap between RL and MPC-1+ is therefore due to routing, not to the release dates; so is the advantage of RL over the classical baselines, whose costs change little with the learned offset.
\item \emph{Why the offset matters little.} Above about $1.0\,T$ the release constraint rarely binds: a job's next batch can start only when its previous batch is complete, and with offsets of $1.0\,T$ or more that date is usually later than the release date (with EST/FIFO in the exact settings, the release date binds for 14--18\% of the job-rounds at $1.0\,T$ and 1--2\% at $1.5\,T$). The learned offsets ($0.94\,T$--$1.30\,T$ on average) thus mostly amount to releasing jobs as early as possible, and the Gaussian release head contributes little; a state-dependent release policy would be needed to exploit release dates.
\item \emph{Per-seed results.} The per-seed reductions against MPC-1+ with the same offset range from $-55.9\%$ ($6\times4$, exact) to $+16.9\%$ ($3\times3$, noisy); for $10\times5$ with exact durations one of the five seeds is better than MPC-1+ ($+14.6\%$) and the others are 10.7\% to 48.2\% worse.
\end{itemize}

\subsubsection{Computational cost: NumPy and Numba}\label{sec:cost}

All experiments were run twice, with the NumPy and with the Numba implementation. The two runs give the same schedules, the same costs for every method, seed, test instance and noise realization, and the same learned policies (the weights agree to within $10^{-7}$, the difference coming from the order of floating-point additions). Table~\ref{tab:compute} compares their running times.

\begin{table}[t]
\centering
\caption{Computational cost of the NumPy and Numba implementations (identical results). Online time per round (policy or controller plus simulation, test episodes, exact durations), training time per seed (300 iterations; mean over the settings with and without noise) and share of the simulator in the time spent inside training and test episodes. $^{e}$~The $3\times3$ shops use $8\times8$ episodes per iteration, the others $6\times6$. $^{f}$~Remaining time: policy decoding and Python bookkeeping.}
\label{tab:compute}
\scriptsize
\setlength{\tabcolsep}{3pt}
\begin{tabular}{l rrr rrr rrr}
\toprule
 & \multicolumn{3}{c}{$3\times3$} & \multicolumn{3}{c}{$6\times4$} & \multicolumn{3}{c}{$10\times5$}\\
\cmidrule(lr){2-4}\cmidrule(lr){5-7}\cmidrule(lr){8-10}
ms per round & NumPy & Numba & ratio & NumPy & Numba & ratio & NumPy & Numba & ratio\\
\midrule
Dispatching rule (EST/FIFO) & 0.30 & 0.038 & 8$\times$ & 0.79 & 0.039 & 20$\times$ & 1.64 & 0.044 & 38$\times$ \\
MPC-1 & 0.68 & 0.103 & 7$\times$ & 7.53 & 1.39 & 5$\times$ & 17.5 & 2.86 & 6$\times$ \\
MPC-1+ ($K=300$) & 79.9 & 7.38 & 11$\times$ & 269 & 7.57 & 36$\times$ & 675 & 9.12 & 74$\times$ \\
RL & 0.33 & 0.047 & 7$\times$ & 0.83 & 0.049 & 17$\times$ & 1.81 & 0.049 & 37$\times$ \\
RL + search, $K=50$ & 13.5 & 1.17 & 12$\times$ & 48.5 & 1.30 & 37$\times$ & 115 & 1.63 & 70$\times$ \\
RL + search, $K=100$ & 26.4 & 2.13 & 12$\times$ & 96.6 & 2.38 & 41$\times$ & 225 & 3.00 & 75$\times$ \\
\midrule
RL training, 300 iterations (s per seed)$^{e}$ & 66\,s & 10.5\,s & 6.2$\times$ & 105\,s & 6.3\,s & 16.7$\times$ & 238\,s & 6.9\,s & 34.6$\times$ \\
Share of the simulator in the episode time$^{f}$ & 27\% & 25\% &  & 18\% & 24\% &  & 15\% & 22\% &  \\
\bottomrule
\end{tabular}
\end{table}

\paragraph{Observations.}
\begin{itemize}[leftmargin=*]
\item \emph{Training} takes 6.3 to 10.5\,s per seed with Numba against 66 to 238\,s with NumPy, 6.2 times faster for $3\times3$, 16.7 times for $6\times4$ and 34.6 times for $10\times5$: the gain grows with the size, as in the simulation benchmark.
\item \emph{Decisions} are 7 to 37 times faster for the policy, and 11 to 75 times faster for the controllers that evaluate many schedules per round (MPC-1+, RL + search); for MPC-1 the gain is only 5 to 7 times because its random candidate sequences are still drawn in NumPy.
\item \emph{The simulator is not the bottleneck.} Inside the episodes the max-plus simulation takes 15--27\% of the time with NumPy and 22--25\% with Numba; the rest is the Giffler--Thompson decoding with the policy and Python bookkeeping. A faster simulator alone would therefore speed up training by at most $1.2\times$--$1.4\times$; the measured gains come from compiling the decoding and the simulator together, which the simple structure of both makes easy. In this application the speed of the max-plus model matters most where it is called many times per decision: in MPC-1+ and in the decision-time search.
\end{itemize}

\section{Discussion and Limitations}\label{sec:disc}

\paragraph{What is supported.} (i) The max-plus simulators reproduce the event-driven and time-driven results exactly and are up to $194\times$ and $419\times$ faster in our benchmark; the factor needs a compiled, sparse implementation. (ii) In the job-shop application, a policy learned on simulated instances outperforms the dispatching rules and a myopic one-round MPC in mean cost in all six settings (significantly in five) and a derivative-free search in the same policy class (significantly in four). A non-myopic one-round MPC (MPC-1+) is as good on the smallest shops and better on the larger ones, but needs 150--190 times more computation per decision. Using the learned policy to propose candidate schedules for the max-plus model at decision time gives the lowest mean cost in four settings and beats MPC-1+ in five, at a third of its decision time. The ablation shows that none of these differences comes from the release offsets. (iii) The structural results (translation equivariance, liveness of sequence-encoded modes) are proven. (iv) A compiled implementation of policy and simulator gives the same results as the NumPy one and trains 6--35 times faster.

\paragraph{What is not shown.} The link between the two parts of the paper is only partly established. The simulation benchmark measures long simulations, whereas in the RL loop the simulator is called one round at a time and takes 15--27\% of the episode time, so the end-to-end gain comes mainly from compiling policy and simulator together; we did not train the same policy with an event-driven environment inside the loop, nor measure RL performance as a function of the simulation budget. The breadth of the approach (other systems than job shops) rests on the structure of the dynamics, not on experiments. The benchmark compares with our own reference simulators (Section~\ref{sec:bench}). The RL results hold for one family of instances, up to $10\times5$ shops, 40 test instances per setting and 5 training seeds. The strongest baseline is a one-round MPC with 300 candidates; a longer-horizon MPC or a mixed-integer formulation could be stronger, and RL + search, like MPC, uses the model at decision time.

\section{Conclusion and Future Work}

Reinforcement learning is limited by simulation cost, and discrete event systems driven by synchronization and delays have an exact, cheap simulator: the max-plus recursion, linear for timed event graphs and switched for systems with conflicts. We formulated control of such systems as an MDP on this recursion, with a relative-date state justified by translation equivariance and a conflict-resolution scheme that always gives live modes. Sparse, compiled max-plus simulators were up to $194\times$ faster than an event-driven simulator and up to $419\times$ faster than a time-driven one on job shops up to $20\times10$, with identical outputs and a 2.6 to 6.0 times smaller memory footprint; the max-plus structure alone, without compilation, gave $2\times$--$9\times$. As an application, a policy trained on such simulations lowered the mean cost of repetitive job-shop scheduling by 24--62\% against dispatching rules and a myopic one-round MPC, and by 18--26\% against a derivative-free search in the same policy class. A non-myopic one-round MPC was stronger than the policy on the larger shops, but 150--190 times slower per decision; letting the policy propose candidate schedules to the max-plus model at decision time beat that MPC in five of six settings at a third of its decision time. A compiled implementation trained 6--35 times faster than NumPy with identical results; in the training loop the simulator itself was a minor share of the time.

We believe that this pairing is a general recipe: for systems whose dynamics are (switched) max-plus linear, the same exact model can serve as training environment and as a fast model for decision-time search. That belief is a conjecture supported by one application. Future work: (i) end-to-end measurement of RL training time and of the learning curve as a function of the simulation budget, with event-driven environments (pure Python and compiled) inside the same training loop, and with an established simulation library as reference; (ii) batched and GPU implementations of the max-plus recursion (e.g.\ vectorized over many instances or candidates) for massively parallel rollouts and search; (iii) other applications of the same recipe: cyclic timetables of rail and metro networks, multi-product processing networks, supply chains and communication pipelines, and standard job-shop benchmarks; (iv) richer (neural or graph-based) policies that exploit the structure of $A_0$, features that describe the effect of a decision on the next round, and state-dependent release dates; (v) longer-horizon MPC and mixed-integer baselines with run-time comparisons; (vi) within-round replanning and stochastic max-plus models; (vii) safe learning with hard time constraints.

\clearpage
\appendix
\section{Implementation Details of the Application}

\begin{itemize}[leftmargin=*]
\item \textbf{Reward and cost.} $w_{\mathrm{late}}=3$, $w_{\mathrm{early}}=1$; reward $=-$ cost of the round; return = reward-to-go; undiscounted.
\item \textbf{Features.} Seven features (Section~\ref{sec:method}), dates relative to $r_n=\min_j\max(u_j,\text{job completion}_j)$ and divided by $T$.
\item \textbf{Training.} Adam ($\beta_1=0.9,\beta_2=0.999$), learning rate $0.05$; $\sigma$ decays by $0.99$ per iteration to a floor of $0.1$; the release scale is $0.5\,T$; 300 iterations, $8\times8$ rollouts per iteration ($3\times3$) or $6\times6$ ($6\times4$ and $10\times5$); 5 seeds per setting; advantages normalized over the batch.
\item \textbf{Baselines.} Release offsets tuned on 12 tuning instances over the grid $\{0, 0.1, 0.25, 0.5, 0.75, 1,$ $1.25, 1.5, 1.75, 2\}$ (in units of $T$; Table~\ref{tab:tuned}). MPC-1+: the schedules of the six rules (the five dispatching rules and the second MWKR variant) and 300 Giffler--Thompson schedules with random priorities $\theta\sim\mathcal{N}(0,25\,I)$ sampled from the softmax, terminal weight $\lambda\in\{0,0.3,1,3\}$ tuned jointly with the offset. CEM: population 24, 6 elite, 6 fresh instances per iteration, initial standard deviation 5 for the weights and 2 for $b$, fitness = mean log cost, as many training episodes as REINFORCE.
\item \textbf{RL + search.} Greedy schedule plus $K$ schedules sampled from the policy, duplicates removed, evaluated on the nominal durations; $(\lambda,\tau_s)$ tuned per seed on the tuning instances (Table~\ref{tab:search_params}).
\item \textbf{Evaluation of the max-plus model.} The matrix form uses $A_0,A_1,B$ and a Kleene-star solver; a vectorized evaluator processes the sequence of operations and was verified to agree with it on random modes and durations before the experiments.
\item \textbf{Implementations.} NumPy (reference) and Numba (compiled round simulator and Giffler--Thompson decoding); a start-up check verifies that both give the same schedules, gradients and random-number state on random cases. Both were run; results are identical (Section~\ref{sec:cost}).
\item \textbf{Code.} The code will be released soon on Github \href{https://github.com/boussad83/maxplus-rl}{here}.
\end{itemize}

\clearpage
\section{Example Instances and Schedules}\label{app:examples}

\begin{figure}[h]
\centering
\includegraphics[width=\textwidth]{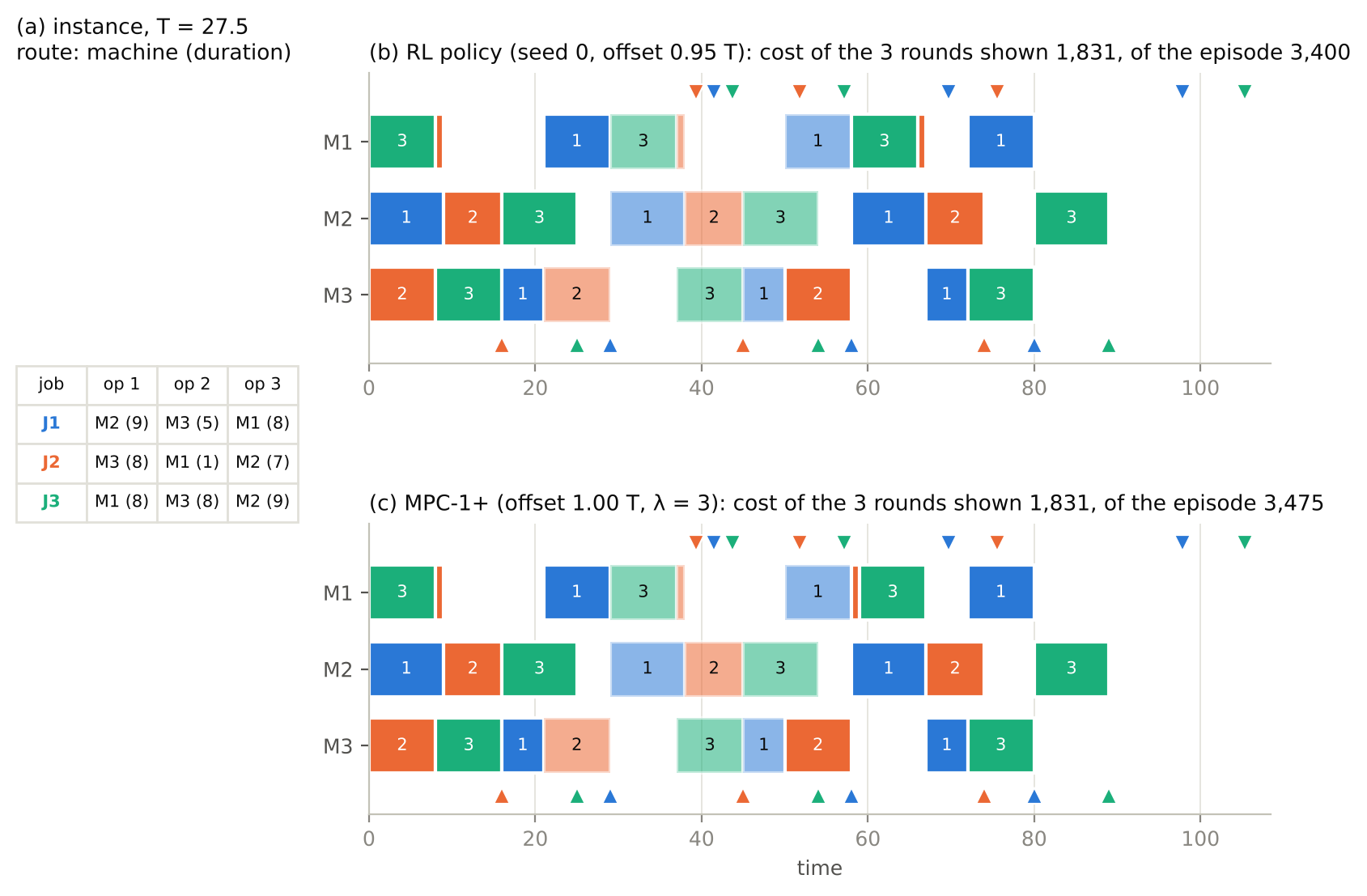}
\caption{First test instance of the $3\times3$ setting with exact durations (period $T=27.5$) and the first three rounds of the schedules of (b) the learned policy (seed 0) and (c) MPC-1+. The two schedules have the same cost over the three rounds shown (1\,831); over the whole episode the policy is slightly better (3\,400 against 3\,475), in line with the near-equality of the two methods on this setting (Table~\ref{tab:mpcp}).}
\label{fig:cfg33}
\end{figure}

\begin{figure}[h]
\centering
\includegraphics[width=\textwidth]{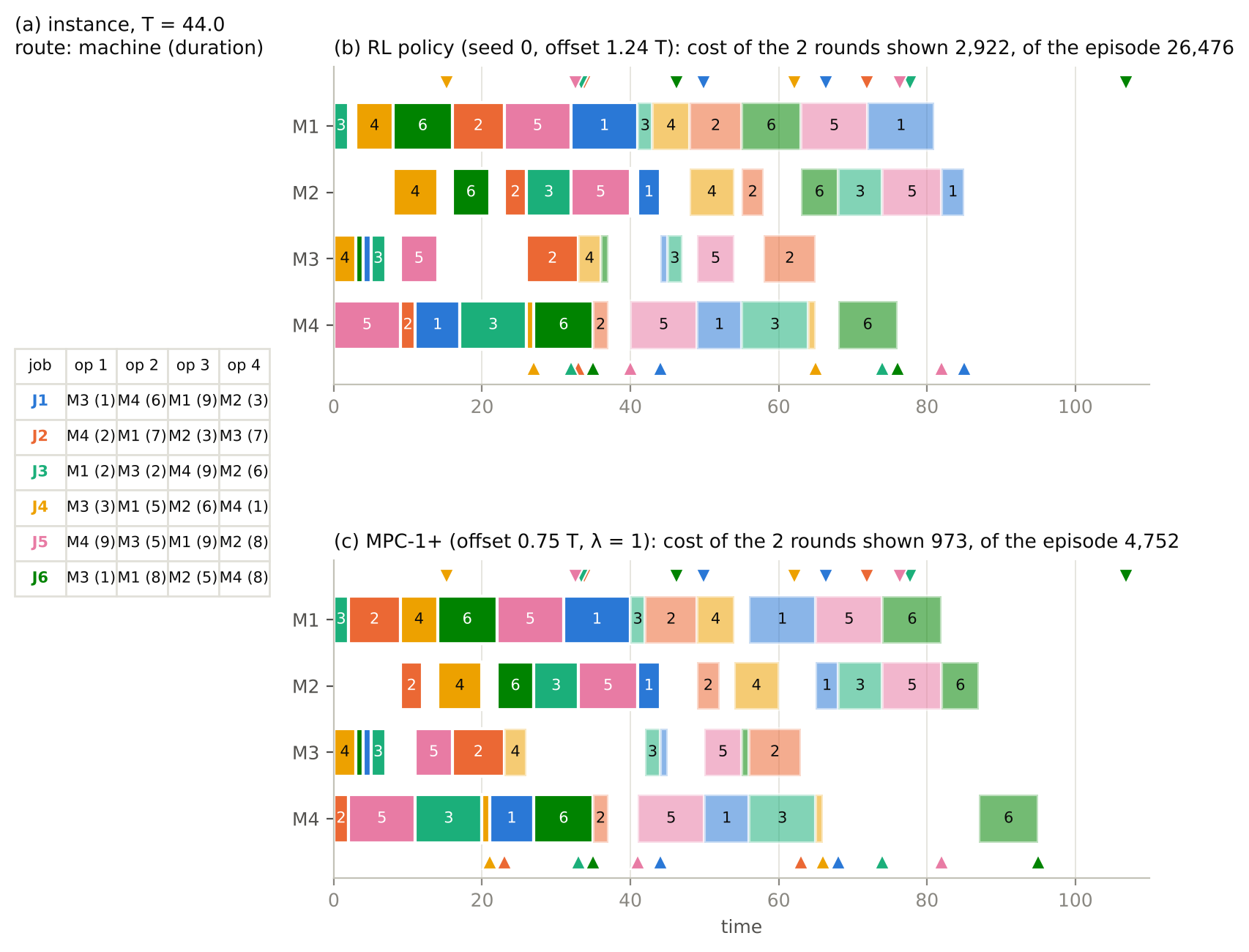}
\caption{First test instance of the $6\times4$ setting with exact durations (period $T=44.0$) and the first two rounds of the schedules of (b) the learned policy (seed 0) and (c) MPC-1+. On this instance MPC-1+ is much better: 973 against 2\,922 over the two rounds shown and 4\,752 against 26\,476 over the episode. This is the first test instance, not a selected one; in this setting the greedy policy beats MPC-1+ on only 15\% of the test instances (Table~\ref{tab:mpcp}).}
\label{fig:cfg64}
\end{figure}

Figures~\ref{fig:cfg33} and~\ref{fig:cfg64} show the first test instance of the $3\times3$ and $6\times4$ settings with exact durations, with the routes and nominal durations, and the schedules of the first rounds produced by the learned policy of seed~0 and by MPC-1+, the best baseline. They are reconstructed from the stored weights and tuned parameters and reproduce the stored test costs of that instance exactly. They illustrate the model and are not statistical evidence. Bars are labelled with the job number; alternating shades correspond to successive rounds; triangles pointing down mark due dates $Z_j(n)$ and triangles pointing up mark completion dates $y_j(n)$, in the colour of the job.

\clearpage
\section{Learning Curves and Test Costs of All Settings}\label{app:allfigs}

\begin{figure}[h]
\centering
\includegraphics[width=0.97\textwidth]{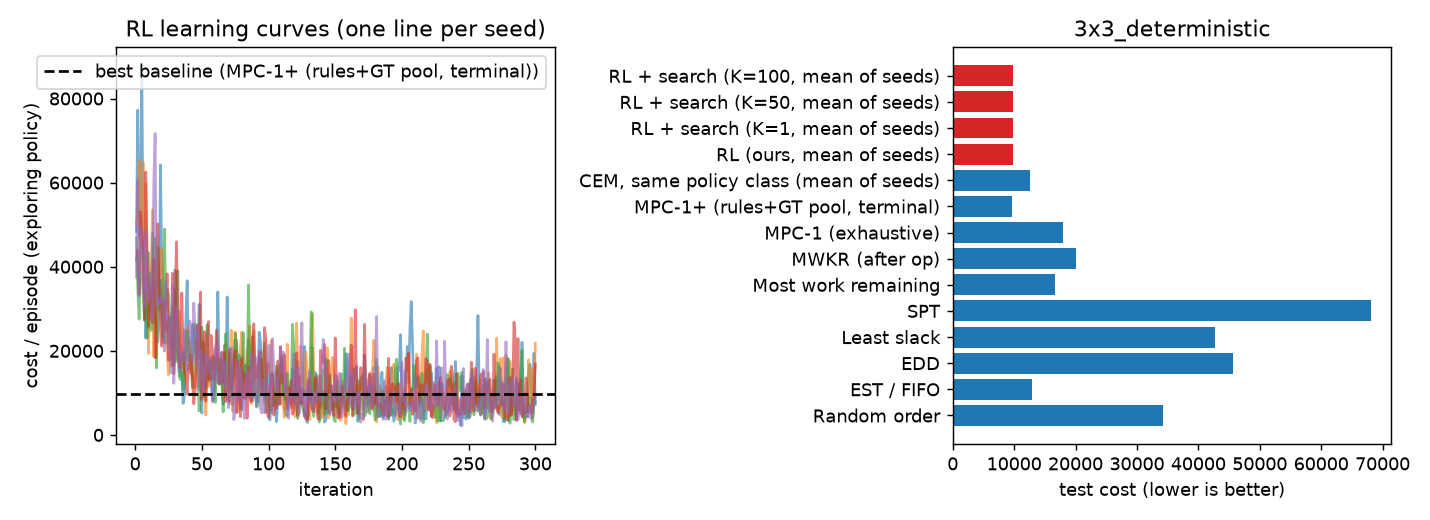}\\[1mm]
\includegraphics[width=0.97\textwidth]{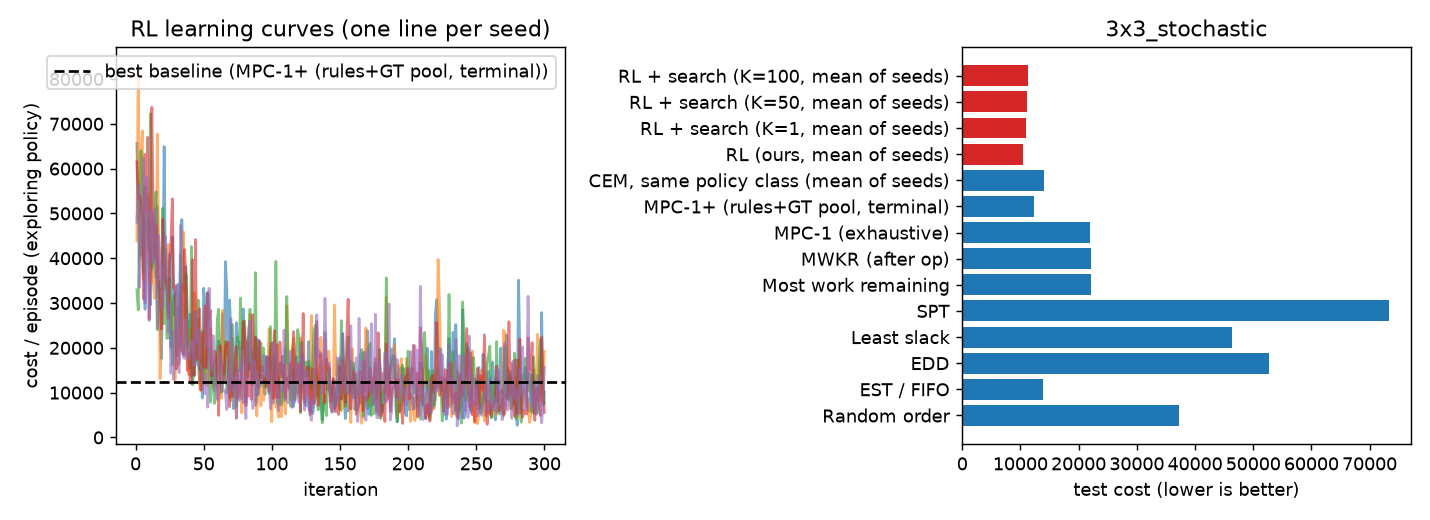}
\caption{$3\times3$ shops: exact durations (top) and $\pm30\%$ noise (bottom).}
\label{fig:a1}
\end{figure}

\begin{figure}[h]
\centering
\includegraphics[width=0.97\textwidth]{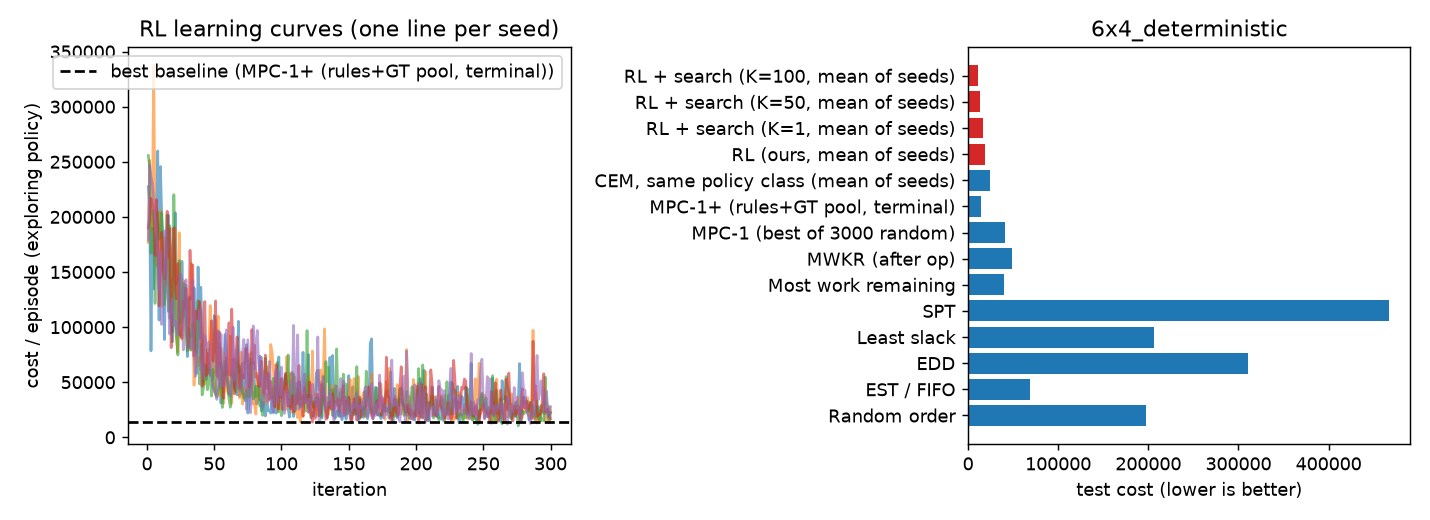}\\[1mm]
\includegraphics[width=0.97\textwidth]{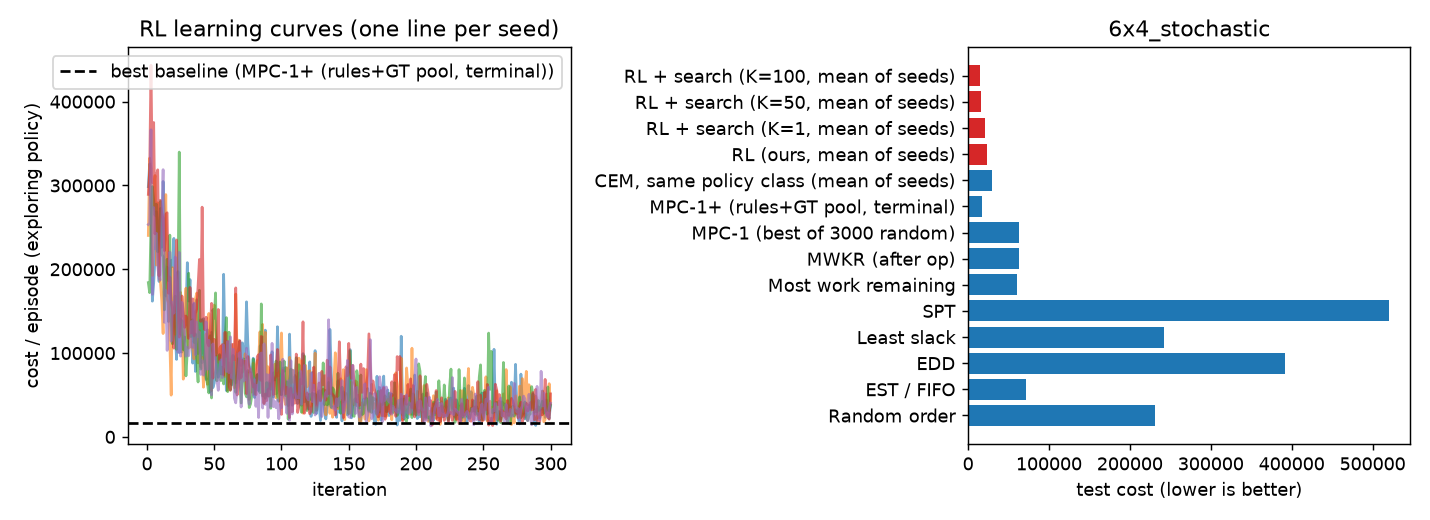}
\caption{$6\times4$ shops: exact durations (top) and $\pm30\%$ noise (bottom).}
\label{fig:a2}
\end{figure}

\begin{figure}[h]
\centering
\includegraphics[width=0.97\textwidth]{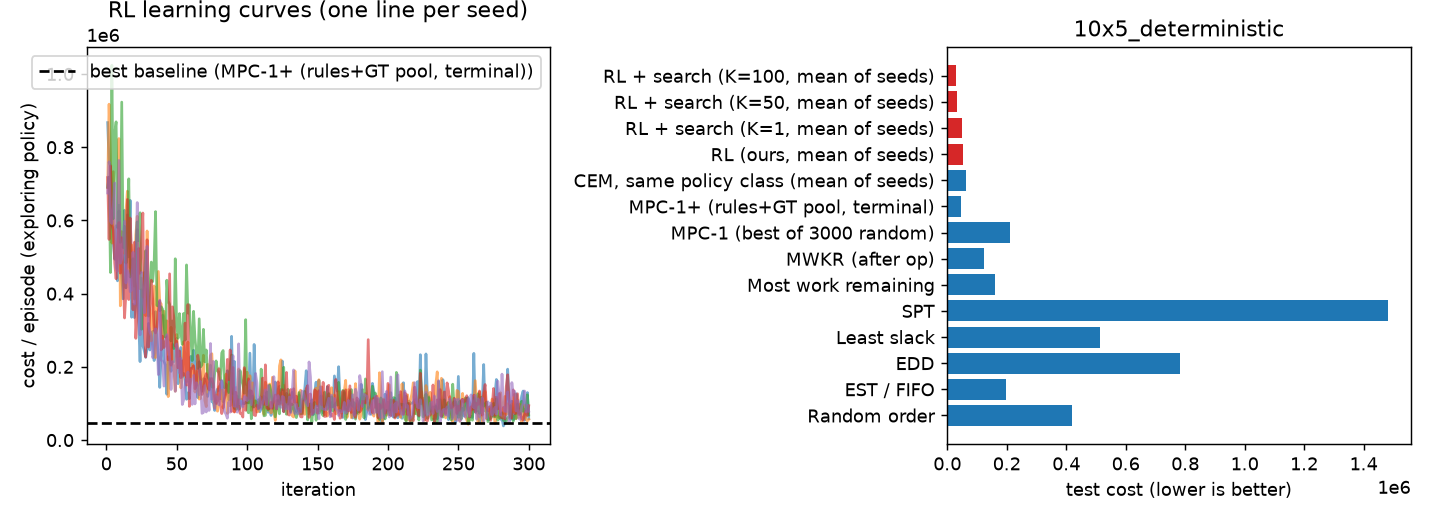}\\[1mm]
\includegraphics[width=0.97\textwidth]{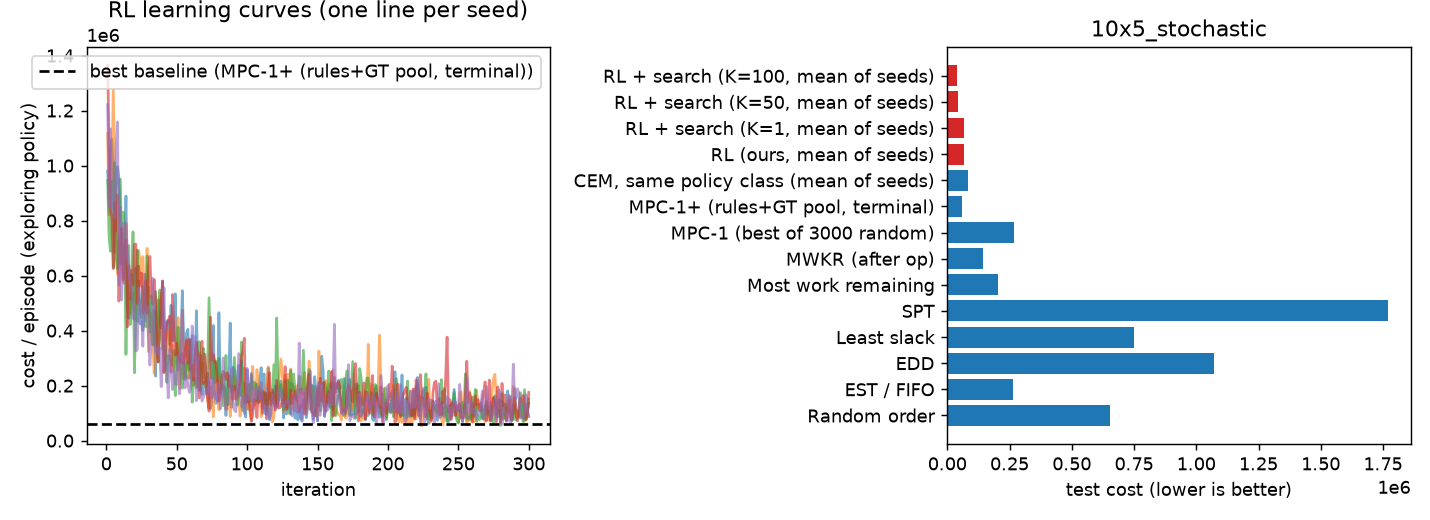}
\caption{$10\times5$ shops: exact durations (top) and $\pm30\%$ noise (bottom).}
\label{fig:a3}
\end{figure}

Figures~\ref{fig:a1}--\ref{fig:a3} show, for each setting, the learning curves of the 5 seeds (left; cost per episode of the exploring policy, dashed line: test cost of the best baseline, MPC-1+) and the test cost of all methods (right; RL, RL + search and CEM are means over seeds; red bars are the RL variants).

\clearpage
\section{Ablation with the Learned Release Offset}\label{app:abl}

\begin{figure}[h]
\centering
\includegraphics[width=0.78\textwidth]{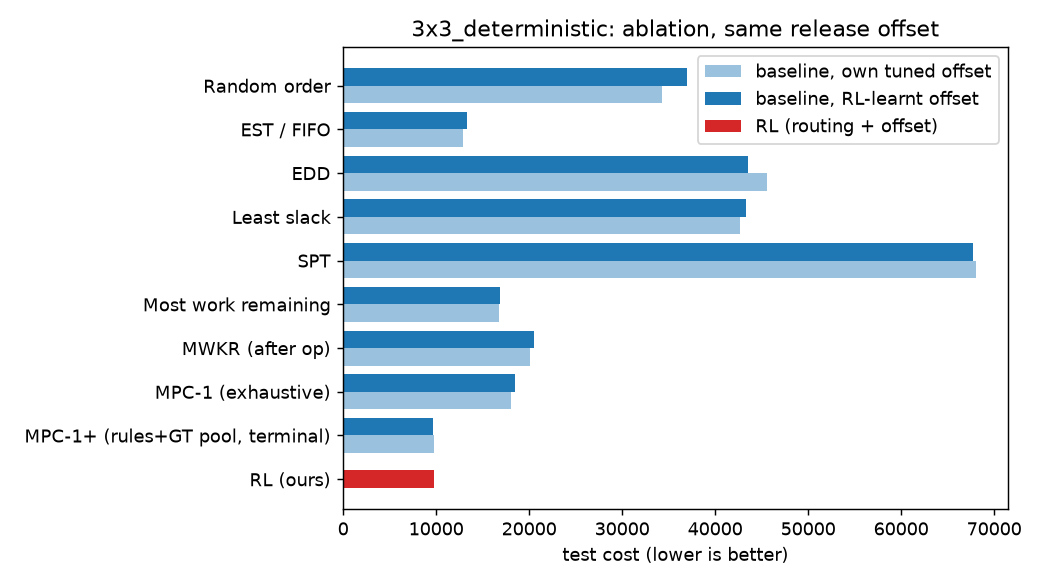}\\[1mm]
\includegraphics[width=0.78\textwidth]{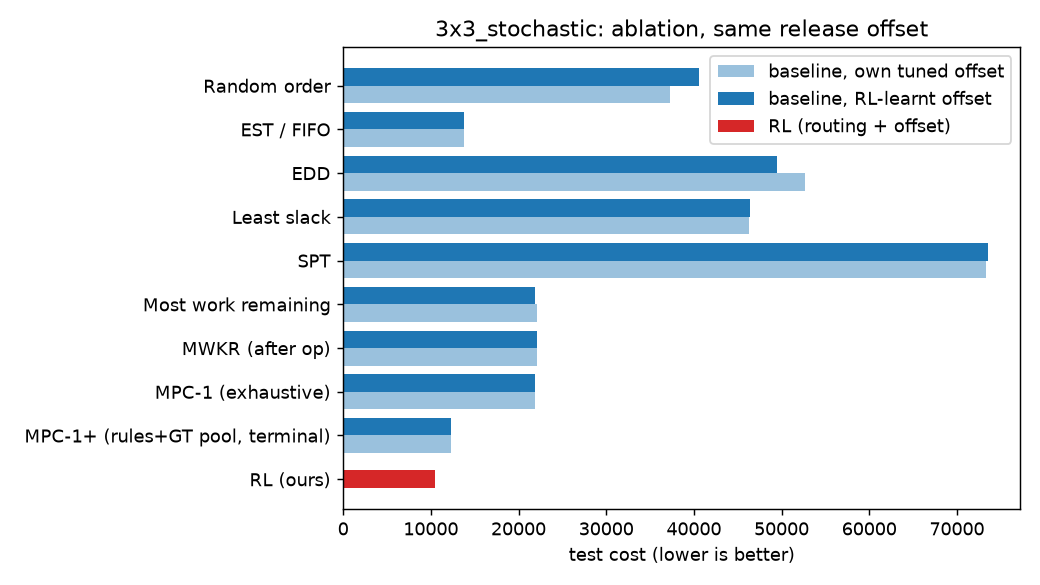}
\caption{Ablation, $3\times3$ shops: exact durations (top) and $\pm30\%$ noise (bottom). Light blue: baselines with their own grid-tuned offset; dark blue: baselines with the offset learned by the policy; red: RL (mean over seeds).}
\label{fig:a4}
\end{figure}

\begin{figure}[h]
\centering
\includegraphics[width=0.78\textwidth]{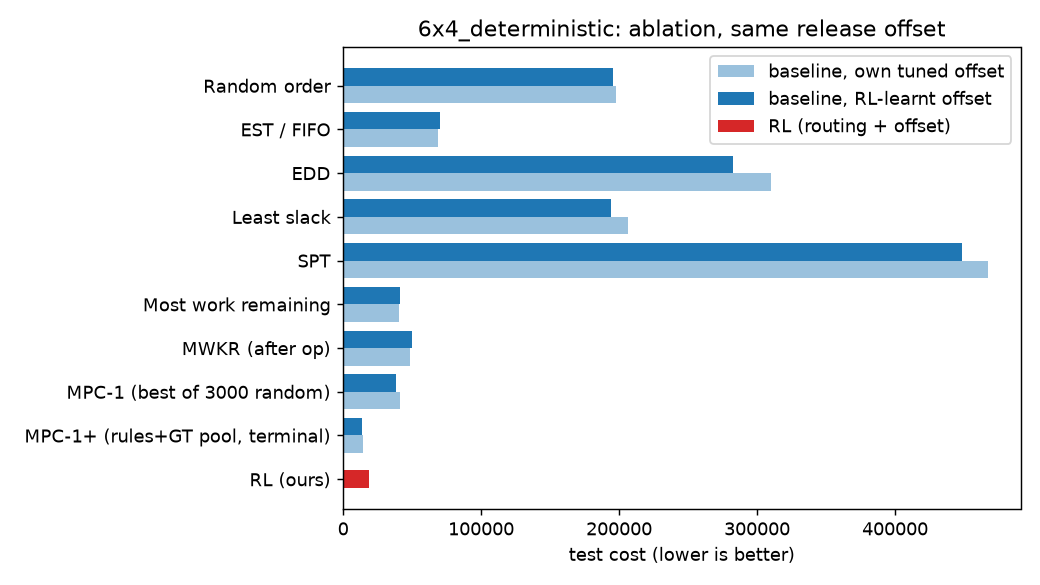}\\[1mm]
\includegraphics[width=0.78\textwidth]{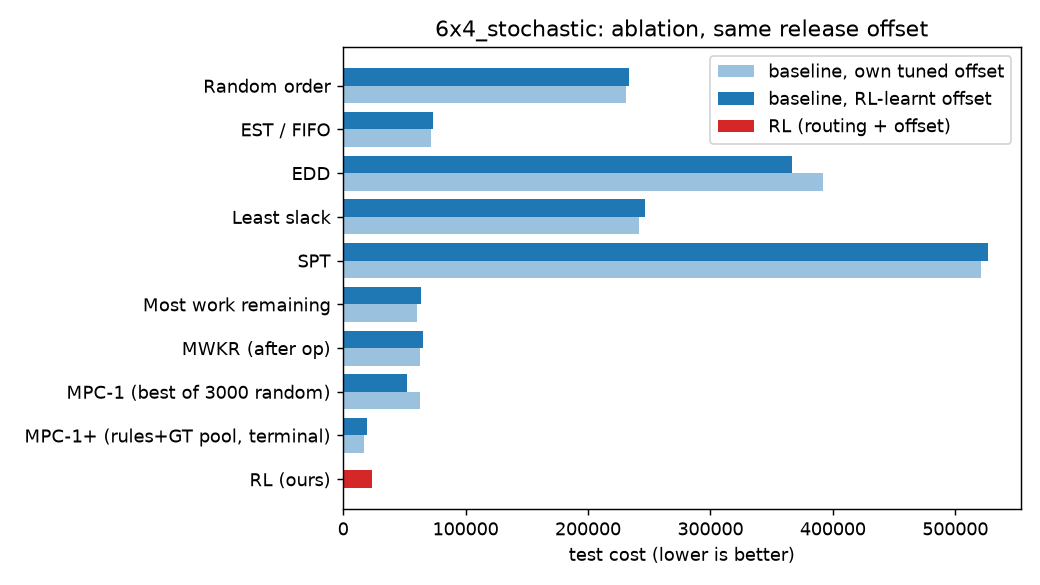}
\caption{Ablation, $6\times4$ shops: exact durations (top) and $\pm30\%$ noise (bottom). Light blue: baselines with their own grid-tuned offset; dark blue: baselines with the offset learned by the policy; red: RL (mean over seeds).}
\label{fig:a5}
\end{figure}

\begin{figure}[h]
\centering
\includegraphics[width=0.78\textwidth]{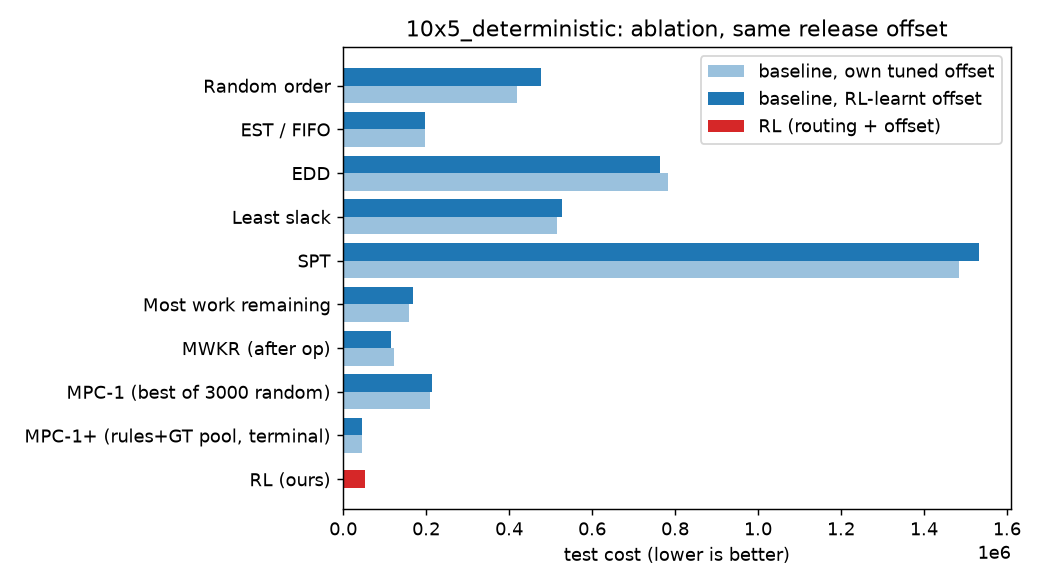}\\[1mm]
\includegraphics[width=0.78\textwidth]{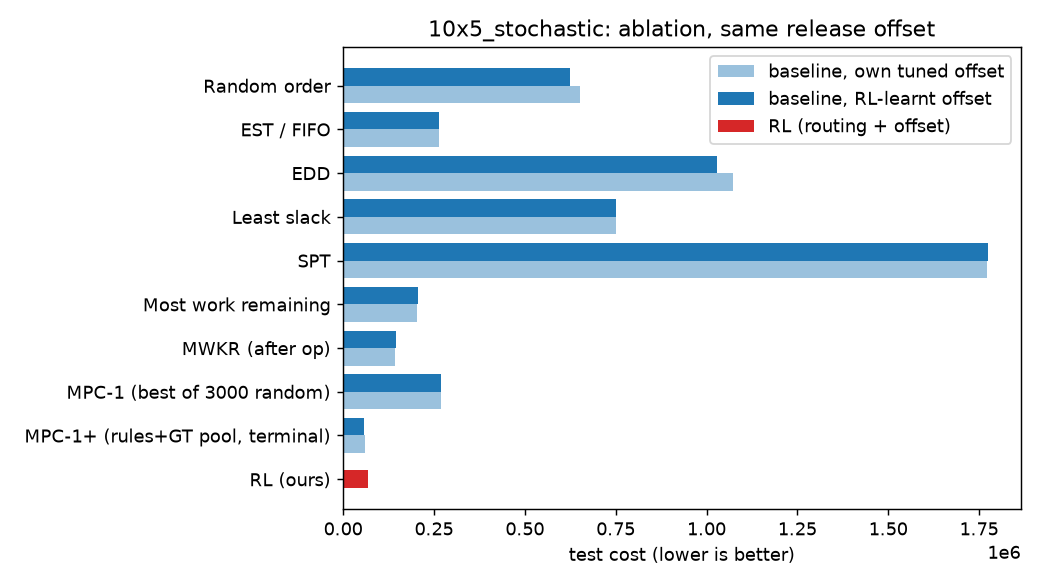}
\caption{Ablation, $10\times5$ shops: exact durations (top) and $\pm30\%$ noise (bottom). Light blue: baselines with their own grid-tuned offset; dark blue: baselines with the offset learned by the policy; red: RL (mean over seeds).}
\label{fig:a6}
\end{figure}

Table~\ref{tab:offsets} gives the release offsets learned by the policy and used for the baselines in the ablation of Section~\ref{sec:abl}. Tables~\ref{tab:abl_exact} and~\ref{tab:abl_noisy} give the cost of all baselines under that offset, and Figures~\ref{fig:a4}--\ref{fig:a6} show the comparison with the baselines' own tuned offsets.

\begin{table}[h]
\centering
\caption{Release offset learned by the policy, as a fraction of the period $T$, over the 5 training seeds. The baselines' grid is $\{0,0.1,0.25,0.5,0.75,1,1.25,1.5,1.75,2\}$.}
\label{tab:offsets}
\scriptsize
\setlength{\tabcolsep}{4pt}
\begin{tabular}{l r r r r}
\toprule
Setting & Mean & Std.\ dev. & Min & Max\\
\midrule
$3\times3$, exact & 0.94 & 0.14 & 0.73 & 1.14 \\
$3\times3$, $\pm30\%$ & 0.95 & 0.05 & 0.87 & 1.02 \\
$6\times4$, exact & 1.24 & 0.34 & 0.84 & 1.84 \\
$6\times4$, $\pm30\%$ & 1.12 & 0.28 & 0.66 & 1.47 \\
$10\times5$, exact & 0.96 & 0.23 & 0.74 & 1.38 \\
$10\times5$, $\pm30\%$ & 1.30 & 0.15 & 1.08 & 1.51 \\
\bottomrule
\end{tabular}
\end{table}

\begin{table}[h]
\centering
\caption{Ablation, exact durations: mean test cost of every baseline run with the release offset learned by the policy (mean over the 5 seeds), and in parentheses the change relative to the baseline's own grid-tuned offset (Table~\ref{tab:exact}); negative is an improvement. RL costs are those of Table~\ref{tab:exact}. Notes as in Table~\ref{tab:exact}.}
\label{tab:abl_exact}
\scriptsize
\setlength{\tabcolsep}{4pt}
\begin{tabular}{l rrr}
\toprule
Method & $3\times3$ & $6\times4$ & $10\times5$\\
\midrule
Random order & 36\,953 (+7.7\%) & 195\,631 ($-$1.1\%) & 477\,692 (+13.9\%) \\
EST / FIFO & 13\,283 (+2.9\%) & 70\,195 (+1.9\%) & 196\,788 ($-$0.5\%) \\
EDD & 43\,546 ($-$4.5\%) & 282\,896 ($-$8.8\%) & 762\,904 ($-$2.5\%) \\
Least slack & 43\,293 (+1.5\%) & 193\,754 ($-$6.1\%) & 528\,053 (+2.6\%) \\
SPT & 67\,781 ($-$0.4\%) & 448\,693 ($-$4.0\%) & 1\,532\,216 (+3.3\%) \\
Most work remaining (MWKR) & 16\,816 (+0.6\%) & 41\,280 (+2.4\%) & 169\,405 (+6.1\%) \\
MWKR, after operation$^{a}$ & 20\,490 (+1.7\%) & 49\,927 (+2.4\%) & 115\,787 ($-$5.2\%) \\
MPC-1$^{b}$ & 18\,510 (+2.6\%) & 37\,997 ($-$7.3\%) & 214\,710 (+2.1\%) \\
MPC-1+$^{c}$ & 9\,625 ($-$1.1\%) & 13\,366 ($-$6.2\%) & 44\,547 ($-$4.1\%) \\
\midrule
\textbf{RL (ours)} & \textbf{9\,807} & \textbf{18\,428} & \textbf{52\,433} \\
\bottomrule
\end{tabular}
\end{table}

\begin{table}[h]
\centering
\caption{Ablation, $\pm30\%$ duration noise; same layout as Table~\ref{tab:abl_exact}, changes relative to Table~\ref{tab:noisy}.}
\label{tab:abl_noisy}
\scriptsize
\setlength{\tabcolsep}{4pt}
\begin{tabular}{l rrr}
\toprule
Method & $3\times3$ & $6\times4$ & $10\times5$\\
\midrule
Random order & 40\,602 (+8.8\%) & 233\,713 (+1.3\%) & 624\,668 ($-$4.3\%) \\
EST / FIFO & 13\,808 ($-$0.1\%) & 73\,676 (+2.2\%) & 264\,802 (+0.3\%) \\
EDD & 49\,459 ($-$6.1\%) & 366\,554 ($-$6.6\%) & 1\,028\,284 ($-$4.1\%) \\
Least slack & 46\,361 (+0.2\%) & 246\,181 (+1.7\%) & 749\,984 (+0.0\%) \\
SPT & 73\,574 (+0.3\%) & 526\,942 (+1.2\%) & 1\,775\,244 (+0.2\%) \\
Most work remaining (MWKR) & 21\,843 ($-$1.1\%) & 63\,657 (+5.2\%) & 205\,875 (+1.9\%) \\
MWKR, after operation$^{a}$ & 22\,108 (+0.1\%) & 64\,808 (+3.7\%) & 145\,893 (+2.4\%) \\
MPC-1$^{b}$ & 21\,834 ($-$0.5\%) & 51\,949 ($-$17.1\%) & 269\,912 (+0.0\%) \\
MPC-1+$^{c}$ & 12\,260 ($-$0.0\%) & 19\,497 (+14.6\%) & 57\,255 ($-$4.7\%) \\
\midrule
\textbf{RL (ours)} & \textbf{10\,486} & \textbf{23\,304} & \textbf{68\,226} \\
\bottomrule
\end{tabular}
\end{table}

\clearpage
\section{Tuned Parameters and Detailed Comparison with MPC-1+}\label{app:param}

\begin{table}[h]
\centering
\caption{Release offsets (fraction of $T$) selected on the tuning instances for the main baselines, and offset / $\lambda$ of MPC-1+.}
\label{tab:tuned}
\scriptsize
\setlength{\tabcolsep}{4pt}
\begin{tabular}{l rrrrr r}
\toprule
Setting & Random & EST/FIFO & MWKR & MWKR, after op. & MPC-1 & MPC-1+\\
\midrule
$3\times3$, exact & 1.00 & 0.75 & 1.00 & 1.00 & 0.50 & 1.00 / 3 \\
$3\times3$, $\pm30\%$ & 1.25 & 1.00 & 1.25 & 1.00 & 0.75 & 1.00 / 1 \\
$6\times4$, exact & 1.00 & 1.00 & 1.00 & 1.00 & 0.50 & 0.75 / 1 \\
$6\times4$, $\pm30\%$ & 1.25 & 1.00 & 1.00 & 1.00 & 0.50 & 0.75 / 3 \\
$10\times5$, exact & 1.25 & 0.75 & 1.00 & 0.75 & 1.00 & 0.75 / 1 \\
$10\times5$, $\pm30\%$ & 1.00 & 1.00 & 1.00 & 1.00 & 1.25 & 1.50 / 3 \\
\bottomrule
\end{tabular}
\end{table}

\begin{table}[h]
\centering
\caption{Parameters $(\lambda,\tau_s)$ of RL + search selected on the tuning instances, for the 5 seeds.}
\label{tab:search_params}
\scriptsize
\setlength{\tabcolsep}{3pt}
\begin{tabular}{l lll}
\toprule
Setting & $K=1$ & $K=50$ & $K=100$\\
\midrule
$3\times3$, exact & (1,1), (3,3), (1,1), (3,3), (3,3) & (1,1), (1,1), (1,1), (1,1), (1,1) & (1,1), (1,1), (1,1), (1,1), (1,1) \\
$3\times3$, $\pm30\%$ & (3,3), (3,3), (3,1), (3,3), (3,3) & (1,1), (1,1), (1,1), (1,1), (1,1) & (1,3), (1,1), (3,1), (1,1), (1,1) \\
$6\times4$, exact & (1,1), (3,3), (1,1), (1,1), (1,3) & (1,3), (1,1), (1,1), (1,1), (1,3) & (1,1), (1,1), (1,1), (1,3), (1,1) \\
$6\times4$, $\pm30\%$ & (0,1), (3,1), (3,1), (0,1), (0,1) & (1,3), (3,3), (1,1), (1,1), (1,1) & (1,3), (3,1), (1,3), (1,1), (1,1) \\
$10\times5$, exact & (0,1), (3,1), (1,1), (3,1), (0,3) & (0,1), (0,1), (1,1), (0,1), (1,1) & (1,3), (1,1), (1,1), (1,1), (1,1) \\
$10\times5$, $\pm30\%$ & (0,3), (0,3), (3,1), (0,3), (0,1) & (1,1), (1,1), (1,1), (0,1), (1,1) & (0,1), (0,1), (1,1), (1,1), (1,1) \\
\bottomrule
\end{tabular}
\end{table}

\begin{table}[h]
\centering
\caption{RL and RL + search versus MPC-1+: relative reduction of the mean cost, mean paired difference (MPC-1+ minus learned controller; positive favours the learned controller) with its bootstrap 95\% interval, share of test instances won / lost, and median of the per-instance cost ratio (learned controller / MPC-1+; below 1 favours the learned controller).}
\label{tab:mpcp_detail}
\scriptsize
\setlength{\tabcolsep}{3pt}
\begin{tabular}{l l r r r r r}
\toprule
Setting & Controller & Reduction & Paired diff. & Bootstrap 95\% CI & Win/loss & Median ratio\\
\midrule
$3\times3$, exact & RL & $-$0.8\% & $-$79 & [$-$2\,071,\ 2\,331] & 50/50\% & 0.99 \\
 & RL + search, $K=50$ & $-$0.8\% & $-$81 & [$-$923,\ 740] & 65/30\% & 0.90 \\
 & RL + search, $K=100$ & $-$0.5\% & $-$50 & [$-$888,\ 770] & 62/30\% & 0.89 \\
\addlinespace
$3\times3$, $\pm30\%$ & RL & +14.5\% & 1\,778 & [$-$794,\ 5\,387] & 38/62\% & 1.16 \\
 & RL + search, $K=50$ & +8.8\% & 1\,079 & [161,\ 2\,594] & 57/42\% & 0.99 \\
 & RL + search, $K=100$ & +8.3\% & 1\,018 & [227,\ 2\,255] & 55/42\% & 0.99 \\
\addlinespace
$6\times4$, exact & RL & $-$29.3\% & $-$4\,177 & [$-$9\,695,\ 3\,298] & 15/85\% & 2.12 \\
 & RL + search, $K=50$ & +10.7\% & 1\,525 & [$-$1\,575,\ 5\,237] & 35/65\% & 1.34 \\
 & RL + search, $K=100$ & +21.0\% & 2\,999 & [$-$1\,042,\ 8\,187] & 38/62\% & 1.29 \\
\addlinespace
$6\times4$, $\pm30\%$ & RL & $-$37.0\% & $-$6\,289 & [$-$10\,456,\ $-$1\,464] & 10/90\% & 1.43 \\
 & RL + search, $K=50$ & +7.9\% & 1\,337 & [$-$23,\ 2\,748] & 65/35\% & 0.94 \\
 & RL + search, $K=100$ & +10.4\% & 1\,770 & [772,\ 2\,901] & 75/25\% & 0.86 \\
\addlinespace
$10\times5$, exact & RL & $-$12.9\% & $-$5\,989 & [$-$19\,326,\ 9\,748] & 25/75\% & 2.06 \\
 & RL + search, $K=50$ & +26.1\% & 12\,133 & [1\,205,\ 25\,182] & 35/65\% & 1.27 \\
 & RL + search, $K=100$ & +34.4\% & 15\,955 & [2\,487,\ 31\,699] & 38/62\% & 1.26 \\
\addlinespace
$10\times5$, $\pm30\%$ & RL & $-$13.6\% & $-$8\,167 & [$-$14\,735,\ $-$1\,752] & 30/70\% & 1.13 \\
 & RL + search, $K=50$ & +28.0\% & 16\,824 & [10\,876,\ 23\,950] & 90/10\% & 0.72 \\
 & RL + search, $K=100$ & +31.6\% & 19\,005 & [12\,658,\ 26\,440] & 90/10\% & 0.69 \\
\bottomrule
\end{tabular}
\end{table}

\clearpage
\section{Mean Total Tardiness}\label{app:tard}

\begin{table}[H]
\centering
\caption{Mean total tardiness per episode (sum over rounds and jobs of the positive part of $y_j-Z_j$); lower is better. Notes as in Table~\ref{tab:exact}.}
\label{tab:tard}
\scriptsize
\setlength{\tabcolsep}{3pt}
\begin{tabular}{l rr rr rr}
\toprule
& \multicolumn{2}{c}{$3\times3$} & \multicolumn{2}{c}{$6\times4$} & \multicolumn{2}{c}{$10\times5$}\\
\cmidrule(lr){2-3}\cmidrule(lr){4-5}\cmidrule(lr){6-7}
Method & exact & $\pm30\%$ & exact & $\pm30\%$ & exact & $\pm30\%$\\
\midrule
Random order & 336.0 & 359.0 & 1\,288.2 & 1\,387.8 & 2\,357.9 & 3\,077.0 \\
EST / FIFO & 133.6 & 139.1 & 489.1 & 561.9 & 1\,293.0 & 1\,544.3 \\
EDD & 383.0 & 428.2 & 1\,648.3 & 1\,894.4 & 3\,533.2 & 4\,212.6 \\
Least slack & 362.8 & 392.3 & 1\,374.8 & 1\,454.9 & 2\,840.1 & 3\,533.1 \\
SPT & 426.7 & 459.7 & 1\,914.3 & 2\,058.0 & 4\,627.8 & 5\,151.4 \\
Most work remaining (MWKR) & 169.2 & 201.8 & 415.3 & 574.8 & 1\,070.9 & 1\,356.9 \\
MWKR, after operation$^{a}$ & 176.7 & 199.5 & 453.8 & 573.0 & 992.1 & 1\,019.8 \\
MPC-1$^{b}$ & 238.0 & 269.9 & 546.2 & 717.3 & 1\,732.0 & 2\,014.9 \\
MPC-1+$^{c}$ & 93.3 & 138.5 & 183.8 & 195.9 & 533.1 & 404.0 \\
CEM, same policy class$^{d}$ & 161.4 & 172.7 & 323.3 & 349.9 & 639.5 & 806.0 \\
\midrule
\textbf{RL (ours)} & 114.2 & 125.7 & 168.2 & 246.6 & 407.7 & 461.3 \\
RL + search, $K=1$ & 112.7 & 125.4 & 152.0 & 245.1 & 406.7 & 524.0 \\
RL + search, $K=50$ & 116.6 & 132.6 & 112.6 & 181.9 & 370.0 & 378.7 \\
RL + search, $K=100$ & 116.5 & 130.0 & 98.3 & 178.1 & 222.0 & 403.4 \\
\bottomrule
\end{tabular}
\end{table}

\clearpage
\section{Variability across Training Seeds and Learned Weights}\label{app:seeds}

\begin{table}[h]
\centering
\caption{Test cost of the RL policy over the 5 independent training seeds (each seed evaluated on the full test sample).}
\label{tab:seeds}
\scriptsize
\setlength{\tabcolsep}{4pt}
\begin{tabular}{l r r r r r r}
\toprule
Setting & Seeds & Mean & Std.\ dev. & Std./mean & Min & Max\\
\midrule
$3\times3$, exact & 5 & 9\,807 & 255 & 2.6\% & 9\,466 & 10\,094 \\
$3\times3$, $\pm30\%$ & 5 & 10\,486 & 257 & 2.5\% & 10\,195 & 10\,881 \\
$6\times4$, exact & 5 & 18\,428 & 1\,089 & 5.9\% & 16\,901 & 19\,568 \\
$6\times4$, $\pm30\%$ & 5 & 23\,304 & 1\,493 & 6.4\% & 21\,071 & 25\,030 \\
$10\times5$, exact & 5 & 52\,433 & 8\,015 & 15.3\% & 46\,572 & 65\,056 \\
$10\times5$, $\pm30\%$ & 5 & 68\,226 & 4\,718 & 6.9\% & 63\,520 & 74\,371 \\
\bottomrule
\end{tabular}
\end{table}

\begin{table}[h]
\centering
\caption{Learned weights $\theta$ (mean $\pm$ standard deviation over the 5 seeds) for the features: earliest start (EST), projected slack, duration, remaining work after the operation, due date, negative part of the slack, and fraction of remaining operations. Features are relative to the reference date and divided by $T$.}
\label{tab:weights}
\scriptsize
\setlength{\tabcolsep}{2.5pt}
\begin{tabular}{l rrrrrrr}
\toprule
Setting & EST & Slack & Duration & Rem.\ work & Due date & Neg.\ slack & Rem.\ ops\\
\midrule
$3\times3$, exact & $-$9.4 $\pm$ 0.2 & 0.4 $\pm$ 0.3 & $-$1.3 $\pm$ 0.3 & 5.6 $\pm$ 0.2 & $-$1.6 $\pm$ 0.2 & 1.0 $\pm$ 0.4 & 4.9 $\pm$ 0.2 \\
$3\times3$, $\pm30\%$ & $-$9.7 $\pm$ 0.2 & 0.9 $\pm$ 0.3 & $-$1.1 $\pm$ 0.4 & 5.4 $\pm$ 0.2 & $-$1.8 $\pm$ 0.1 & 1.0 $\pm$ 0.4 & 5.0 $\pm$ 0.1 \\
$6\times4$, exact & $-$10.6 $\pm$ 0.4 & $-$1.5 $\pm$ 0.4 & $-$1.8 $\pm$ 0.8 & 5.6 $\pm$ 0.3 & $-$1.0 $\pm$ 0.3 & 0.0 $\pm$ 0.5 & 5.3 $\pm$ 0.3 \\
$6\times4$, $\pm30\%$ & $-$11.4 $\pm$ 0.2 & $-$1.0 $\pm$ 0.1 & $-$1.9 $\pm$ 0.2 & 5.8 $\pm$ 0.3 & $-$0.8 $\pm$ 0.2 & $-$0.2 $\pm$ 0.3 & 6.0 $\pm$ 0.1 \\
$10\times5$, exact & $-$11.3 $\pm$ 0.4 & $-$2.1 $\pm$ 0.4 & $-$2.4 $\pm$ 0.7 & 6.0 $\pm$ 0.4 & $-$1.0 $\pm$ 0.4 & $-$0.7 $\pm$ 0.5 & 5.8 $\pm$ 0.2 \\
$10\times5$, $\pm30\%$ & $-$11.5 $\pm$ 0.3 & $-$1.4 $\pm$ 0.2 & $-$2.9 $\pm$ 0.5 & 5.9 $\pm$ 0.2 & $-$0.4 $\pm$ 0.4 & $-$0.7 $\pm$ 0.2 & 6.0 $\pm$ 0.4 \\
\bottomrule
\end{tabular}
\end{table}

\clearpage
\section{Full Simulators Benchmark Results}\label{app:bench}

All numbers come from two independent benchmark runs; in the first the event-driven simulator is the reference of the speedup column, in the second the time-driven one. ``ok'' was true for every row: all simulators returned the same makespan for the same instance. Times are wall-clock; the CPU time (not shown) is within a few percent of the wall time, except for a few very short runs.

\begin{table}[h]
\centering
\caption{Wall-clock time per simulation. Event-driven and time-driven times are those of their own benchmark run; the max-plus times are those of the first run.}
\scriptsize
\setlength{\tabcolsep}{4pt}
\begin{tabular}{l r rr rrr}
\toprule
Shop & $K$ & event & time-driven & NumPy dense & Numba dense & Numba sparse\\
\midrule
$3\times3$ & $10^2$ & 1.05 ms & 1.84 ms & 0.75 ms & 0.13 ms & 0.13 ms \\
$3\times3$ & $10^3$ & 10.34 ms & 18.93 ms & 5.59 ms & 0.17 ms & 0.19 ms \\
$3\times3$ & $10^4$ & 106.19 ms & 190.12 ms & 57.36 ms & 0.98 ms & 0.88 ms \\
\addlinespace
$6\times4$ & $10^2$ & 2.78 ms & 4.09 ms & 1.18 ms & 0.36 ms & 0.19 ms \\
$6\times4$ & $10^3$ & 28.65 ms & 41.27 ms & 8.72 ms & 0.60 ms & 0.26 ms \\
$6\times4$ & $10^4$ & 290.97 ms & 418.39 ms & 82.26 ms & 4.69 ms & 2.02 ms \\
\addlinespace
$10\times5$ & $10^2$ & 5.62 ms & 11.32 ms & 1.82 ms & 0.58 ms & 0.27 ms \\
$10\times5$ & $10^3$ & 56.31 ms & 119.45 ms & 13.31 ms & 2.17 ms & 0.53 ms \\
$10\times5$ & $10^4$ & 613.00 ms & 1.14 s & 136.25 ms & 19.65 ms & 4.12 ms \\
\addlinespace
$20\times10$ & $10^2$ & 21.98 ms & 47.04 ms & 8.34 ms & 4.81 ms & 0.63 ms \\
$20\times10$ & $10^3$ & 236.45 ms & 502.20 ms & 62.82 ms & 38.06 ms & 2.18 ms \\
$20\times10$ & $10^4$ & 2.40 s & 5.00 s & 630.70 ms & 386.22 ms & 12.36 ms \\
\bottomrule
\end{tabular}
\end{table}

\begin{table}[h]
\centering
\caption{Speedup of the max-plus implementations over the event-driven simulator (left) and over the time-driven simulator with $dt=1$ (right), excluding JIT compilation. Each speedup is computed within its own benchmark run.}
\scriptsize
\setlength{\tabcolsep}{4pt}
\begin{tabular}{l r rrr rrr}
\toprule
 & & \multicolumn{3}{c}{vs event-driven} & \multicolumn{3}{c}{vs time-driven}\\
\cmidrule(lr){3-5}\cmidrule(lr){6-8}
Shop & $K$ & NumPy & Numba d. & Numba s. & NumPy & Numba d. & Numba s.\\
\midrule
$3\times3$ & $10^2$ & 1.4 & 7.9 & 8.2 & 2.4 & 15 & 16 \\
$3\times3$ & $10^3$ & 1.9 & 60 & 54 & 3.3 & 85 & 88 \\
$3\times3$ & $10^4$ & 1.9 & 108 & 120 & 3.5 & 196 & 198 \\
\addlinespace
$6\times4$ & $10^2$ & 2.4 & 7.8 & 14 & 3.9 & 13 & 20 \\
$6\times4$ & $10^3$ & 3.3 & 47 & 112 & 4.8 & 65 & 162 \\
$6\times4$ & $10^4$ & 3.5 & 62 & 144 & 5.0 & 89 & 207 \\
\addlinespace
$10\times5$ & $10^2$ & 3.1 & 9.7 & 21 & 6.4 & 18 & 52 \\
$10\times5$ & $10^3$ & 4.2 & 26 & 107 & 8.6 & 57 & 220 \\
$10\times5$ & $10^4$ & 4.5 & 31 & 149 & 8.8 & 58 & 282 \\
\addlinespace
$20\times10$ & $10^2$ & 2.6 & 4.6 & 35 & 5.3 & 9.7 & 72 \\
$20\times10$ & $10^3$ & 3.8 & 6.2 & 108 & 7.2 & 13 & 232 \\
$20\times10$ & $10^4$ & 3.8 & 6.2 & 194 & 8.6 & 14 & 419 \\
\bottomrule
\end{tabular}
\end{table}

\begin{table}[h]
\centering
\caption{Peak Python heap (MB) for the five simulators, and peak resident-set-size increase (MB) for the event-driven, time-driven and Numba sparse simulators.}
\scriptsize
\setlength{\tabcolsep}{3pt}
\begin{tabular}{l r rrrrr rrr}
\toprule
 & & \multicolumn{5}{c}{Peak heap [MB]} & \multicolumn{3}{c}{Peak RSS increase [MB]}\\
\cmidrule(lr){3-7}\cmidrule(lr){8-10}
Shop & $K$ & event & time & NumPy & Numba d. & Numba s. & event & time & Numba s.\\
\midrule
$3\times3$ & $10^2$ & 0.03 & 0.03 & 0.01 & 0.01 & 0.01 & 0.18 & 0.48 & 0.14 \\
$3\times3$ & $10^3$ & 0.32 & 0.39 & 0.08 & 0.07 & 0.08 & 0.49 & 0.88 & 0.21 \\
$3\times3$ & $10^4$ & 3.15 & 4.12 & 0.69 & 0.69 & 0.69 & 3.52 & 4.61 & 0.83 \\
\addlinespace
$6\times4$ & $10^2$ & 0.06 & 0.06 & 0.04 & 0.03 & 0.03 & 0.20 & 0.51 & 0.18 \\
$6\times4$ & $10^3$ & 0.49 & 0.69 & 0.21 & 0.20 & 0.20 & 0.65 & 1.21 & 0.34 \\
$6\times4$ & $10^4$ & 4.77 & 7.23 & 1.85 & 1.84 & 1.84 & 5.32 & 8.18 & 1.99 \\
\addlinespace
$10\times5$ & $10^2$ & 0.14 & 0.16 & 0.12 & 0.07 & 0.06 & 0.29 & 0.62 & 0.20 \\
$10\times5$ & $10^3$ & 1.40 & 1.70 & 0.46 & 0.41 & 0.40 & 1.64 & 2.24 & 0.54 \\
$10\times5$ & $10^4$ & 14.86 & 18.38 & 3.89 & 3.85 & 3.84 & 15.84 & 19.82 & 3.98 \\
\addlinespace
$20\times10$ & $10^2$ & 0.39 & 0.43 & 0.92 & 0.70 & 0.23 & 0.55 & 0.91 & 0.33 \\
$20\times10$ & $10^3$ & 3.90 & 4.48 & 2.29 & 1.89 & 1.60 & 4.23 & 4.96 & 1.71 \\
$20\times10$ & $10^4$ & 39.91 & 46.98 & 16.02 & 15.62 & 15.33 & 45.66 & 52.54 & 15.44 \\
\bottomrule
\end{tabular}
\end{table}

\begin{table}[h]
\centering
\caption{Number of simulations after which the one-off JIT compilation time of the Numba sparse simulator (about 0.13--0.15\,s) is recovered with respect to the event-driven simulator, i.e.\ JIT time divided by the time saved per simulation. Values below 1 mean that a single simulation already pays for the compilation.}
\label{tab:benchbet}
\scriptsize
\begin{tabular}{l r r}
\toprule
Shop & $K$ & Break-even (number of runs)\\
\midrule
$3\times3$ & $10^2$ & 158.7 \\
$3\times3$ & $10^3$ & 14.2 \\
$3\times3$ & $10^4$ & 1.3 \\
\addlinespace
$6\times4$ & $10^2$ & 51.9 \\
$6\times4$ & $10^3$ & 4.8 \\
$6\times4$ & $10^4$ & 0.5 \\
\addlinespace
$10\times5$ & $10^2$ & 24.8 \\
$10\times5$ & $10^3$ & 2.5 \\
$10\times5$ & $10^4$ & 0.2 \\
\addlinespace
$20\times10$ & $10^2$ & 6.9 \\
$20\times10$ & $10^3$ & 0.6 \\
$20\times10$ & $10^4$ & 0.1 \\
\bottomrule
\end{tabular}
\end{table}

\clearpage
\section{A photo of a handwritten draft from 2019}
\begin{figure}[h]
\centering
\includegraphics[width=\textwidth]{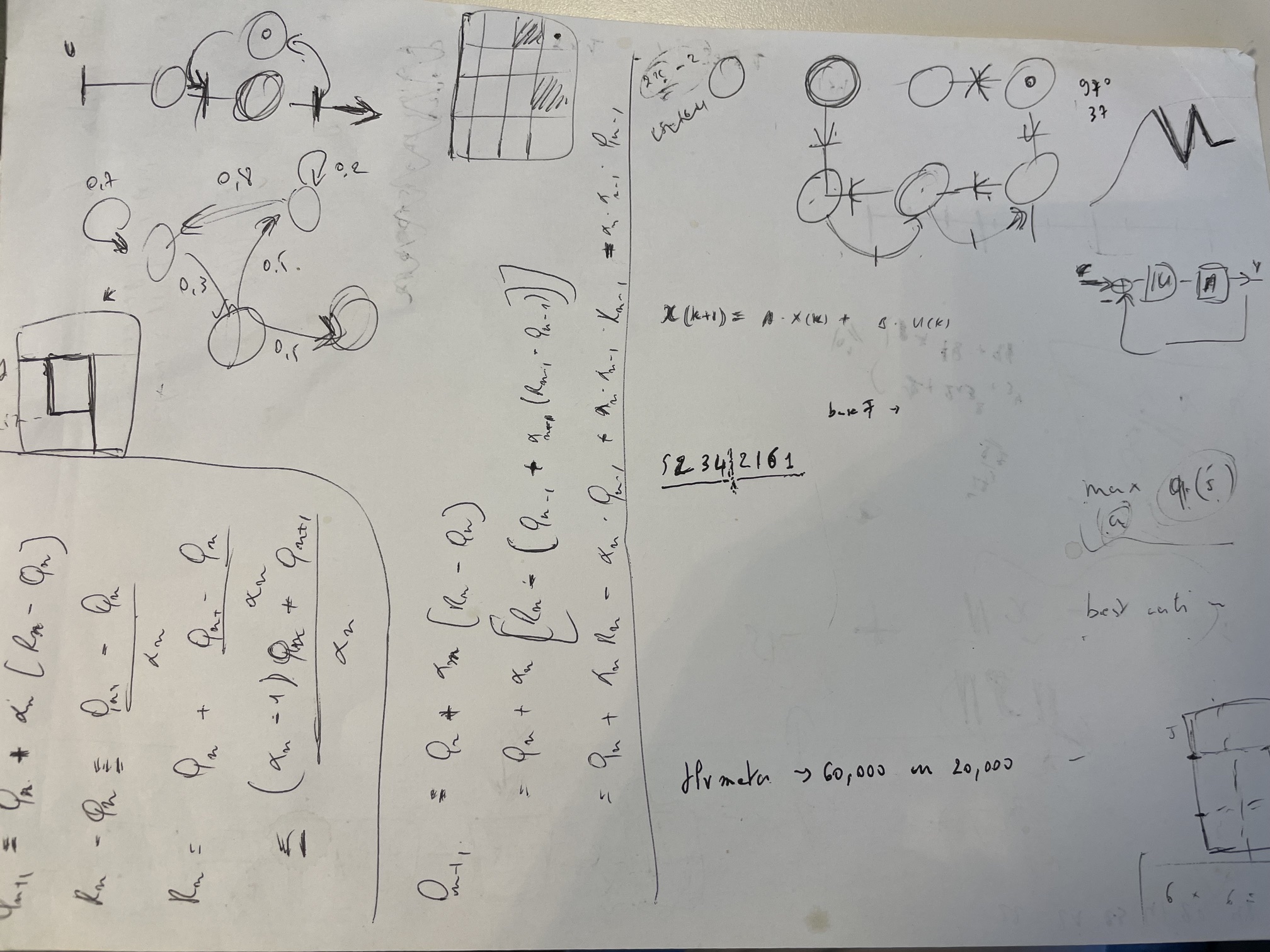}
\caption{The draft shows a timed event graph (TEG) together with its canonical max-plus representation, along with several equations related to reinforcement learning.}
\label{fig:draft2019}
\end{figure}

\end{document}